\documentclass[11pt]{article}

\usepackage[final]{acl}
\usepackage[english]{babel}
\usepackage[utf8]{inputenc}
\usepackage{latexsym}
\usepackage{enumitem}
\usepackage{newtxtext,newtxmath}

\usepackage[T1]{fontenc}
\usepackage{xspace}
\usepackage{xcolor}
\usepackage{amsmath}
\usepackage{cleveref}

\usepackage[utf8]{inputenc}

\usepackage{microtype}

\usepackage{inconsolata}

\usepackage{graphicx}
\usepackage[table]{xcolor}
\usepackage{tabularx}
\usepackage{float}
\usepackage{xltabular}
\usepackage{booktabs}
\usepackage{makecell}
\usepackage{multirow}
\usepackage{threeparttable}

\usepackage{pdfpages}
\usepackage{cleveref}
\usepackage{caption}
\usepackage{subcaption} %
\usepackage{booktabs,multirow}
\usepackage{xeCJK}
\usepackage{soul}
\usepackage{makecell}

\newcommand{\datasetname}{\textcolor{black}{\textsc{DimABSA}}\xspace}

\sethlcolor{yellow}

\title{DimABSA: Building Multilingual and Multidomain Datasets for Dimensional Aspect-Based Sentiment Analysis}

\author{
\textbf{Lung-Hao Lee}$^{1,*}$, \textbf{Liang-Chih Yu}$^{2,*}$, \textbf{Natalia Loukachevitch}$^{3}$, \textbf{Ilseyar Alimova}$^{4}$\\
\textbf{Alexander Panchenko}$^{4,5}$, \textbf{Tzu-Mi Lin}$^{1}$, \textbf{Zhe-Yu Xu}$^{1}$, \textbf{Jian-Yu Zhou}$^{1}$,
\textbf{Guangmin Zheng}$^{6}$\\
\textbf{Jin Wang}$^{6}$, \textbf{Sharanya Awasthi}$^{7}$,  \textbf{Jonas Becker}$^{8}$,
\textbf{Jan Philip Wahle}$^{8}$,
\textbf{Terry Ruas}$^{8}$\\ 
\textbf{Shamsuddeen Hassan Muhammad}$^{9}$, \textbf{Saif M. Mohammad}$^{10}$ \vspace{+0.5em}
\\
$^{1}$National Yang Ming Chiao Tung University, $^{2}$Yuan Ze University \\
$^{3}$Moscow State University, 
$^{4}$Skoltech,
$^{5}$AIRI,
$^{6}$Yunnan University, 
$^{7}$University of Cincinnati \\
$^{8}$University of G\"ottingen,
$^{9}$Imperial College London, 
$^{10}$National Research Council Canada\vspace{+0.5em}
\\
\textbf{*Contact:} \texttt{lhlee@nycu.edu.tw, lcyu@saturn.yzu.edu.tw}
}
\begin{document}
\selectlanguage{english}

\maketitle
\begin{abstract}

Aspect-Based Sentiment Analysis (ABSA) focuses on extracting sentiment at a fine-grained aspect level and has been widely applied across real-world domains. However, existing ABSA research relies on coarse-grained categorical labels (e.g., positive, negative), which limits its ability to capture nuanced affective states. To address this limitation, we adopt a dimensional approach that represents sentiment with continuous valence–arousal (VA) scores, enabling fine-grained analysis at both the aspect and sentiment levels. To this end, we introduce \datasetname, the first multilingual, dimensional ABSA resource annotated with both traditional ABSA elements (aspect terms, aspect categories, and opinion terms) and newly introduced VA scores. This resource contains 76,958 aspect instances across 42,590 sentences, spanning six languages and four domains. We further introduce three subtasks that combine VA scores with different ABSA elements, providing a bridge from traditional ABSA to dimensional ABSA.
Given that these subtasks involve both categorical and continuous outputs, we propose a new unified metric, continuous F1 (cF1), which incorporates VA prediction error into standard F1. We provide a comprehensive benchmark using both prompted and fine-tuned large language models across all subtasks. Our results show that DimABSA is a challenging benchmark and provides a foundation for advancing multilingual dimensional ABSA. We publicly released the \datasetname{} dataset, which was used for Track A of SemEval-2026 Task 3, attracting over 300 participants.\footnote{\url{https://github.com/DimABSA/DimABSA2026}}

\end{abstract}

\section{Introduction}
Aspect-Based Sentiment Analysis (ABSA) aims to analyze opinion structures at the fine-grained aspect level, rather than at the sentence or document level \cite{pontiki-etal-2014-semeval,pontiki-etal-2015-semeval,pontiki2016semeval}. It is formulated as the individual or joint extraction of sentiment elements, including aspect terms, aspect categories, opinion terms, and sentiment polarity. For example, given the sentence \textit{I am happy with the laptop.}, an ABSA system is expected to extract the explicit aspect term \textit{laptop}, the opinion term \textit{happy}, assign the aspect category {\small \texttt{LAPTOP\#GENERAL}} from a predefined set, and predict {\small \texttt{Positive}} sentiment polarity. These elements can form different structural combinations (e.g., aspect pairs, triplets, and quadruplets), enabling fine-grained analysis that provides deeper insights into user opinions across various applications. \citep{d2022knowmis, Zhang2023, hua2024systematic}.

\begin{figure}[t]
    \centering
        \includegraphics[width=\columnwidth]{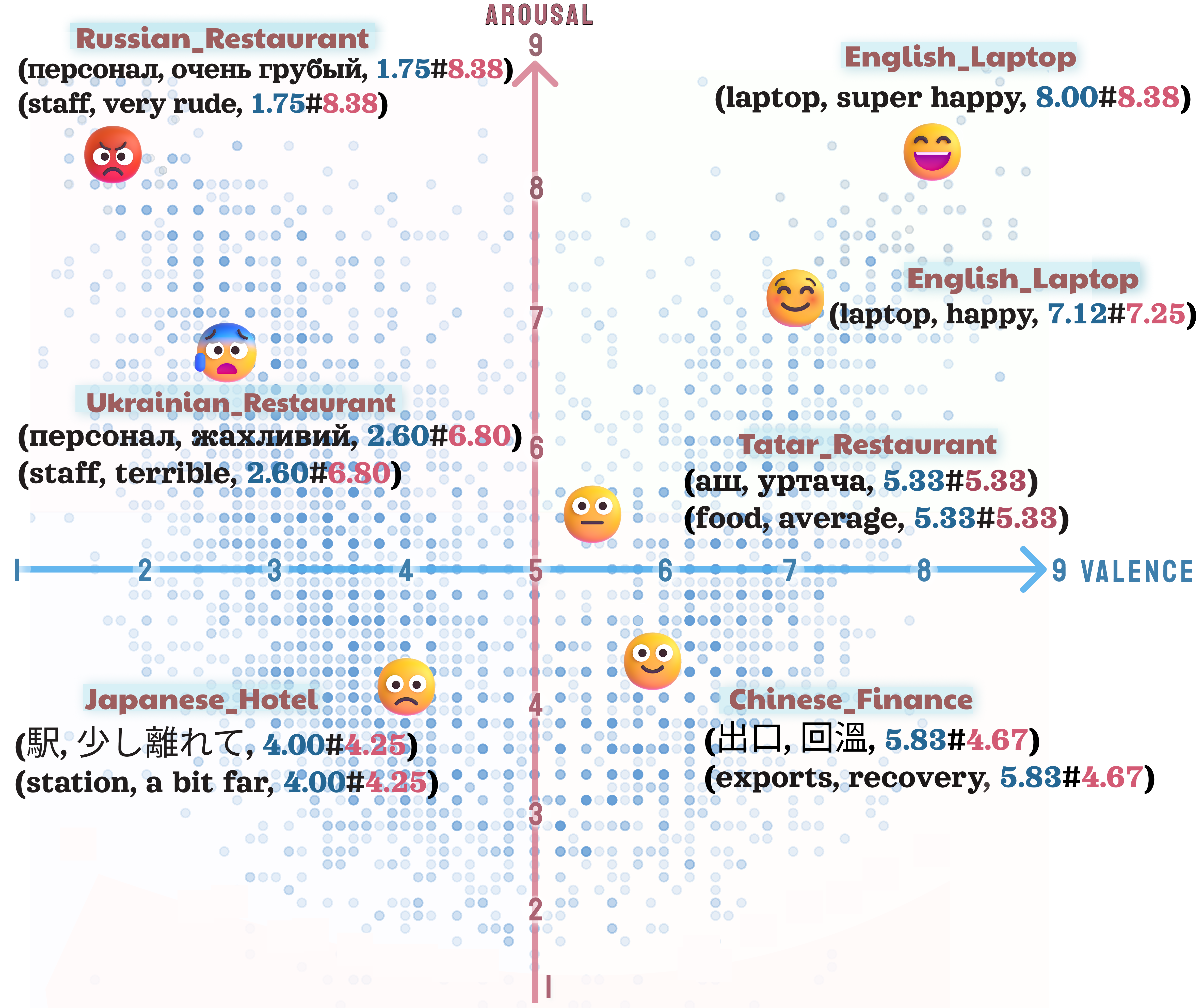}
\vspace{-20pt}
\caption{\textbf{Valence–Arousal (VA) space}. Illustrated examples of aspect instances across multiple languages and domains. Selected instances are visualized as triplets (aspect, opinion, paired VA score). Blue dots show a U-shaped distribution of VA scores.}
\label{fig:VAspace}

\end{figure}

Despite substantial progress in ABSA, existing work remains limited by its reliance on coarse-grained, categorical sentiment labels (e.g., positive, negative, neutral). This categorical approach fails to capture subtle affective differences, such as those conveyed by variations in lexical intensity (e.g., \textit{good}, \textit{excellent}) or by sentiment-modifying adverbs (e.g., \textit{slightly}, \textit{very}).

To address these limitations, we draw inspiration from affective science, which provides nuanced sentiment representation frameworks \citep{russell1980circumplex, russell2003core} that model sentiment along real-valued dimensions of valence (negative–positive) and arousal (sluggish–excited). As illustrated by the aspect instances in \Cref{fig:VAspace}, this dimensional representation captures diverse affective states across languages and domains, rather than being limited to categorical labels. Moreover, even affective states that share the same polarity (e.g., \textit{super happy} vs.\@ \textit{recovery}, and \textit{very rude} vs \textit{a bit far}) can be distinguished through their fine-grained valence–arousal (VA) scores. This expressive capability further extends ABSA to enable fine-grained analysis at both the aspect and sentiment levels. The VA framework has been used to support various applications, such as stance detection \cite{Upadhyaya2023}, misinformation identification \cite{Liu2024}, identifying biosocial markers for mental health \cite{teodorescu-etal-2023-language},
dialogue generation \cite{Wen2024}, hate speech detection \cite{sariyanto-etal-2025-explainable}, and emotion dynamics analysis \cite{pone.0256153,teodorescu-etal-2023-language,wang-etal-2025-feel}.

To move beyond the categorical sentiment representation, we introduce the \datasetname, the first multilingual dimensional ABSA resource manually annotated with both traditional ABSA annotations (aspect terms, aspect categories, opinion terms) and newly introduced VA scores. \datasetname comprises 76,958 aspect instances across 42,590 sentences, covering six languages (Chinese, English, Japanese, Russian, Tatar, and Ukrainian) and four domains, including customer reviews (Hotel, Laptop, Restaurant) and financial reporting (Finance).

\begin{table}[!tb]
\centering
\small
\resizebox{\linewidth}{!}{%
\begin{tabular}{lcc}
\toprule
\textbf{Subtask (Input $\rightarrow$ Output)} & \textbf{Prediction Type} & \textbf{Metric} \\
\midrule
\makecell[l]{\textbf{DimASR}\\(text + aspects $\rightarrow$ V\#A)} 
& Regression & RMSE \\
\makecell[l]{\textbf{DimASTE}\\(text $\rightarrow$ A, O, V\#A)} 
& Extraction, Regression & cF1 \\
\makecell[l]{\textbf{DimASQP}\\(text $\rightarrow$ A, C, O, V\#A)} 
& Extraction, Classification, Regression & cF1 \\ 
\bottomrule
\end{tabular}}
\caption{\textbf{DimABSA subtasks} with input–output structure, task type, and metrics.}
\label{tab:dimabsa_task_summary}
\end{table}

To explore the utility of dimensional sentiment in ABSA, we introduce three novel subtasks by integrating continuous VA scores with traditional ABSA components, each corresponding to distinct application scenarios, as shown in \Cref{tab:dimabsa_task_summary}.

\begin{itemize}[leftmargin=*, itemsep=1pt, topsep=3pt]
\item \textbf{Subtask 1 - Dimensional Aspect Sentiment Regression (DimASR):} A regression task that predicts VA scores for each aspect in a sentence.

\item \textbf{Subtask 2 - Dimensional Aspect Sentiment Triplet Extraction (DimASTE):} A hybrid task that jointly extracts aspect (A) and opinion (O) terms and predicts their associated VA scores.

\item \textbf{Subtask 3 - Dimensional Aspect Sentiment Quadruplet Prediction (DimASQP):} An extension of DimASTE that additionally includes aspect category (C) classification, thus combining extraction, classification, and regression.

\end{itemize}

As DimASTE and DimASQP are hybrid tasks requiring both categorical prediction and VA regression, the standard F1 score, widely used in ABSA, is insufficient to jointly assess these components. Thus, we propose a unified metric: the \textit{continuous F1 (cF1)} score, which incorporates VA prediction error into the F1 formulation. 
Experiments on the DimABSA subtasks with various large language models (LLMs) show that closed-source LLMs provide data-efficient baselines, while fine-tuned large-scale LLMs (70B and 120B) further improve performance. Both model families show performance variability across languages and domains, and still face challenges, particularly for low-resource languages.

Our contributions are summarized as follows: 
\begin{itemize} [itemsep=1pt, topsep=3pt]
    \item We introduce the DimABSA datasets with continuous VA scores, enabling fine-grained analysis across languages and domains.
    \item We design three subtasks to facilitate the transition from categorical to dimensional ABSA.
    \item We propose the cF1 score, a unified metric that integrates categorical and continuous evaluation into a single measure. 
\end{itemize}

\noindent The DimABSA datasets were used for Track A of SemEval-2026 Task 3, attracting over 300 participants \cite{yu-etal-2026-semeval}.

\section{The DimABSA Construction}
We collect data from six languages and four domains. Each instance is annotated with ABSA components and VA ratings, following the specifications of the three target subtasks: DimASR, \mbox{DimASTE}, and DimASQP.

\begin{table*}[t]
\centering
\small
\setlength{\tabcolsep}{3pt}
\renewcommand{\arraystretch}{1.2}
\setlength{\extrarowheight}{1.5pt}

\resizebox{1.0\linewidth}{!}{%
\begin{tabular}{l c c
                cccc
                cc}
\toprule
\multirow{2}{*}{\textbf{Dataset}} & \multirow{2}{*}{\textbf{Source(s)}}& \multirow{2}{*}{\textbf{Subtask}}
& \textbf{Train}
& \textbf{Dev}
& \textbf{Test}
& \textbf{Total}
& \textbf{F1}
& \multirow{2}{*}{\textbf{RMSE (V\#A)}} \\
&&
& \textbf{sent./tuple}
& \textbf{sent./tuple}
& \textbf{sent./tuple}
& \textbf{sent./tuple} 
& \textbf{(A) / (A,C) / (A,C,O)} 
& \\
\midrule

\multirow{2}{*}{eng-rest} & ACOS & ST1
& \multirow{2}{*}{2284 / 3659} & 200 / 340 & 1000 / 1504 & 3484 / 5503 & \multirow{2}{*}{0.892 / 0.778 / 0.703}& \multirow{2}{*}{1.542\#1.883} \\
 & Yelp Open Dataset & ST2–3 &  & 200 / 408 & 1000 / 2129 & 3484 / 6196 &  \\ \hline

\multirow{2}{*}{eng-lap}  & ACOS & ST1
& \multirow{2}{*}{4076 / 5773}  & 200 / 275 & 1000 / 1421 & 5276 / 7469 & \multirow{2}{*}{0.857 / 0.767 / 0.635}& \multirow{2}{*}{1.547\#2.291} \\
 & Amazon Reviews 2023 & ST2–3
&  & 200 / 317 & 1000 / 1975 & 5276 / 8065 & \\ \hline

\multirow{2}{*}{jpn-hot}& \multirow{2}{*}{Rakuten Travel} & ST1
& \multirow{2}{*}{1600 / 2846} & 200 / 284 & 800 / 1092 & 2600 / 4222 & \multirow{2}{*}{0.824 / 0.754 / 0.643}& \multirow{2}{*}{0.919\#1.018} \\
& & ST2–3 & & 200 / 364 & 800 / 1443 & 2600 / 4653 & \\ \hline

jpn-fin & chABSA; EDINET & ST1
& 1024 / 1672 & 200 / 319 & 800 / 1302 & 2024 / 3293 & 0.761 / -- / -- & 0.814\#0.755 \\ \hline

\multirow{2}{*}{rus-rest} & \multirow{2}{*}{SemEval\textquoteright 16} & ST1
& \multirow{2}{*}{1240 / 2487} & 56 / 81 & 1072 / 1637 & 2368 / 4205 & \multirow{2}{*}{0.865 / 0.748 / 0.726}& \multirow{2}{*}{1.030\#2.041} \\
&& ST2–3 && 48 / 102 & 630 / 1310 & 1918 / 3899 &\\ \hline

\multirow{2}{*}{tat-rest} & \multirow{2}{*}{SemEval\textquoteright 16 (MT)} & ST1
& \multirow{2}{*}{1240 / 2487} & 56 / 81 & 1072 / 1637 & 2368 / 4205 & \multirow{2}{*}{0.865 / 0.748 / 0.726}& \multirow{2}{*}{1.030\#2.041} \\
&& ST2–3&& 48 / 102 & 630 / 1310 & 1918 / 3899 &\\ \hline

\multirow{2}{*}{ukr-rest} & \multirow{2}{*}{SemEval\textquoteright 16 (MT)} & ST1
&\multirow{2}{*}{1240 / 2487} & 56 / 81 & 1072 / 1637 & 2368 / 4205 & \multirow{2}{*}{0.865 / 0.748 / 0.726}& \multirow{2}{*}{1.030\#2.041} \\
 &  & ST2–3& & 48 / 102 & 630 / 1310 & 1918 / 3899 &\\ \hline

\multirow{2}{*}{zho-rest} & SIGHAN-2024& ST1
& \multirow{2}{*}{6050 / 8523}  & 225 / 416 & 1000 / 1929 & 7275 / 10868 & \multirow{2}{*}{0.826 / 0.713 / 0.650}& \multirow{2}{*}{0.591\#0.758}  \\
& Google Reviews; PTT & ST2–3
& & 300 / 761 & 1000 / 2861 & 7350 / 12145 & \\ \hline

\multirow{2}{*}{zho-lap} & \multirow{2}{*}{Mobile01} & ST1
& \multirow{2}{*}{3490 / 6502}  & 261 / 431 & 1000 / 2662 & 4751 / 9595 & \multirow{2}{*}{0.803 / 0.723 / 0.602}& \multirow{2}{*}{0.827\#1.074}  \\
&  & ST2–3 & & 300 / 551 & 1000 / 2798 & 4790 / 9851 &  \\ \hline

zho-fin & MOPS & ST1
& 1000 / 2633 & 200 / 563 & 842 / 2354 & 2042 / 5550 & 0.604 / -- / -- & 0.910\#0.810 \\
\bottomrule
\end{tabular}}
\caption{\textbf{DimABSA dataset overview.} 
For each dataset (language–domain), we report the corresponding source(s), subtask type (ST1 vs.\ ST2–3), and the numbers of sentences and tuples in the train/dev/test splits. Tuple agreement is evaluated using F1, and valence–arousal agreement is evaluated using RMSE. Dataset examples are provided in Appendix A.}
\label{tab:dataset_stats_aclstyle}
\end{table*}

\subsection{Data Collection}
As shown in \Cref{tab:dataset_stats_aclstyle}, we used four widely studied ABSA domains: restaurant (\texttt{rest}), laptop (\texttt{lap}), hotel (\texttt{hot}), and finance (\texttt{fin}). For these domains, we consider four high-resource languages: English (\texttt{eng}), Japanese (\texttt{jpn}), Russian (\texttt{rus}), and Chinese (\texttt{zho}). To extend coverage to under-resourced languages, we translated the Russian data into Tatar (\texttt{tar}) and Ukrainian (\texttt{ukr}), leveraging their linguistic and cultural relatedness. While this approach has inherent limitations, it enables the inclusion of languages that would otherwise be unavailable. All translations were reviewed by native speakers and manually corrected to ensure the quality of ABSA-relevant structural annotations.

We collect data from multiple sources, including existing labeled ABSA datasets and newly curated unlabeled data. The existing labeled datasets are used solely for training, while the newly curated data are annotated and split into training, development, and test sets. The data sources for each language are described below.

\textbf{English.} We use the training split of the ACOS dataset \cite{Cai2021}, manually annotating the restaurant and laptop quadruplets in this split with VA scores to replace the original sentiment polarity labels. 
For the development and test sets, we collect restaurant reviews from Yelp Open Dataset\footnote{\href{https://business.yelp.com/data/resources/open-dataset}{https://business.yelp.com/data/resources/open-dataset}} and laptop reviews from Amazon Reviews 2023 \cite{hou2024bridging}.  

\textbf{Japanese.} For the finance domain, the training set is sampled from the chABSA dataset.\footnote{\href{https://github.com/chakki-works/chABSA-dataset}{https://github.com/chakki-works/chABSA-dataset}} 
We manually annotate VA scores for each aspect in these samples, replacing the original sentiment polarity labels. The development and test sets are collected from the same EDINET\footnote{\href{https://disclosure2.edinet-fsa.go.jp}{https://disclosure2.edinet-fsa.go.jp}} sources as chABSA, with samples involving the same companies removed to avoid overlap. For the hotel domain, we crawl reviews from Rakuten Travel.\footnote{\href{https://travel.rakuten.co.jp}{https://travel.rakuten.co.jp}}

\textbf{Russian.} The SemEval-2016 restaurant review dataset \cite{pontiki2016semeval} serves as the data source. The labeled portion contains annotated aspects, their categories, and sentiment polarity. We annotate opinion terms and VA values manually. The unlabeled portion of reviews is used for the development and test sets. 

\textbf{Tatar.} We automatically translate the Russian dataset into Tatar using Yandex Translate. The resulting translations are reviewed by a native speaker. Manual corrections were applied to 45.5\% of the translated instances to assure data quality.

\textbf{Ukrainian.} Similar to Tatar, we translate the Russian dataset into Ukrainian and adopt the same review procedure, with 35.6\% of the instances corrected by native speakers.

\textbf{Chinese.} For the restaurant domain, we use the SIGHAN-2024 dataset \cite{Lee2024} for training, and construct the development and test sets from Google Reviews\footnote{\href{https://customerreviews.google.com}{https://customerreviews.google.com}} and the PTT platform\footnote{\href{https://www.pttweb.cc}{https://www.pttweb.cc}}. For the laptop domain, we crawl reviews from Mobile01\footnote{\href{https://www.mobile01.com/category.php?id=2}{https://www.mobile01.com/category.php?id=2}}. For the finance domain, we collect annual reports of Taiwanese companies from MOPS\footnote{\href{https://emops.twse.com.tw/server-java/t58query}{https://emops.twse.com.tw/server-java/t58query}}.  

We preprocess all collected data by removing duplicate entries, invisible characters, and corrupted encodings. The cleaned texts are then segmented into sentences and prepared for annotation.

\subsection{Annotation Process}
The DimABSA datasets comprise four core components: three categorical ABSA elements and a paired valence--arousal (VA) score. Each component is defined as follows:

\textbf{Aspect Term (A)}: a word or phrase indicating an opinion target, such as \textit{service}, \textit{screen}, \textit{profit}. 

\textbf{Aspect Category (C)}: a predefined Entity\#Attribute label associated with an aspect term (e.g., {\small \texttt{FOOD\#QUALITY}}, {\small \texttt{SERVICE\#GENERAL}}) \cite{pontiki-etal-2015-semeval, pontiki2016semeval}. %

\textbf{Opinion Term (O)}: a sentiment-bearing word or phrase associated with a specific aspect term. The annotation further includes sentiment modifiers to support fine-grained sentiment representation (e.g., \textit{very good}, \textit{extremely bad}, \textit{a little slow}).

\textbf{Valence-Arousal (VA)}: Both the valence and arousal are rated on a 1–9 scale, where 1 denotes extreme negative valence or low arousal, 9 denotes extreme positive valence or high arousal, and 5 denotes neutral valence or medium arousal.

The annotated elements vary across datasets depending on the subtask setting. For finance-domain datasets focused on DimASR, we annotate \texttt{(A, VA)} pairs. For all other datasets, we annotate full \texttt{(A, C, O, VA)} quadruplets. For existing datasets, we supplement the annotations with VA scores and any missing components of the quadruplet. We construct a shared training set for all subtasks. However, we do not use a single shared development/test set across subtasks, since DimASR assumes the aspect term is given as input. Instead, we create a dedicated development/test set for DimASR, and a shared set for DimASTE and DimASQP.

The annotation process is conducted in two phases. We first extract the categorical triplet \texttt{(A, C, O)} for each quadruplet from sentences, followed by the assignment of VA scores for tuples. For tuple extraction, annotators follow guidelines with examples (see Appendix~\ref{sec:annotation-guidelines}), including extracting all valid tuples, incorporating sentiment modifiers, and ensuring exact span matching. Each sentence is annotated independently by two annotators. If both annotators agree, the tuple is accepted. Otherwise, a third annotator adjudicates the disagreement. Instances without consensus among the three annotators are discarded.

The accepted tuples are then annotated with VA ratings. Annotators are introduced to the VA dimensions with illustrative examples (see Appendix~\ref{sec:annotation-guidelines}). The annotation interface is implemented using the Self-Assessment Manikin (SAM) scale \cite{Bradley1994}, accompanied by VA emojis \cite{Kutsuzawa2022} for reference. This pictorial protocol helps annotators assign VA ratings more accurately. Each tuple is annotated by five annotators and the final VA rating is obtained by discarding outlier ratings beyond the mean $\pm$ 1.5 standard deviations and averaging the remaining scores.

\begin{figure*}[ht]
    \centering

    \begin{subfigure}[b]{0.48\textwidth}
        \centering
        \includegraphics[width=\textwidth]{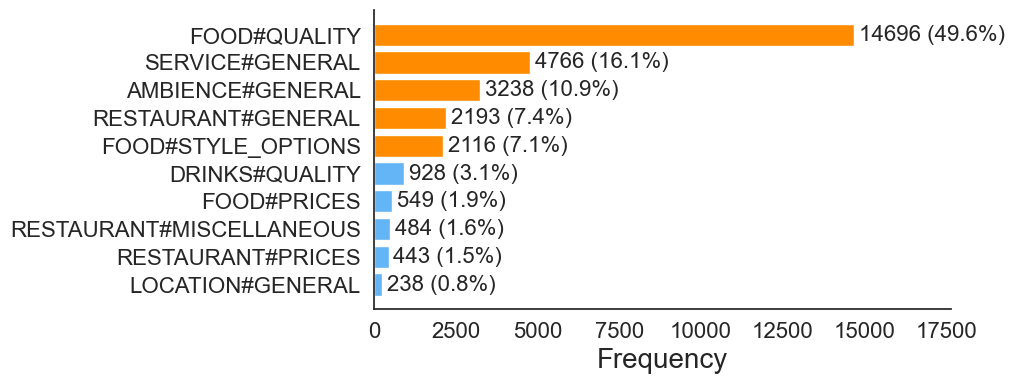}
        \caption{Restaurant}
        \label{fig:sub1}
    \end{subfigure}
    \hfill
    \begin{subfigure}[b]{0.48\textwidth}
        \centering
        \includegraphics[width=\textwidth]
        {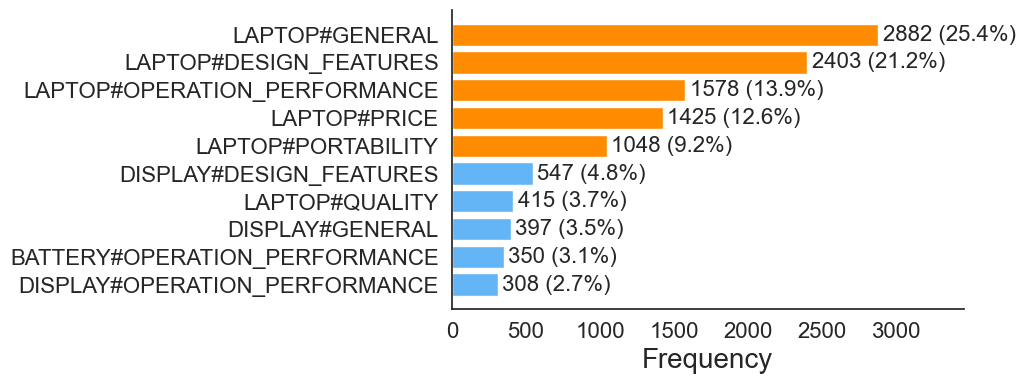}
        \caption{Laptop}
        \label{fig:sub2}
    \end{subfigure}

\vspace{10pt}
\caption{\textbf{Aspect-Category distributions} for the restaurant and laptop domains aggregated across languages.}
\label{fig:category_scatter}
\end{figure*}

\begin{figure*}[ht]
    \centering

    \begin{subfigure}[b]{0.19\textwidth}
        \centering
        \includegraphics[width=\textwidth]{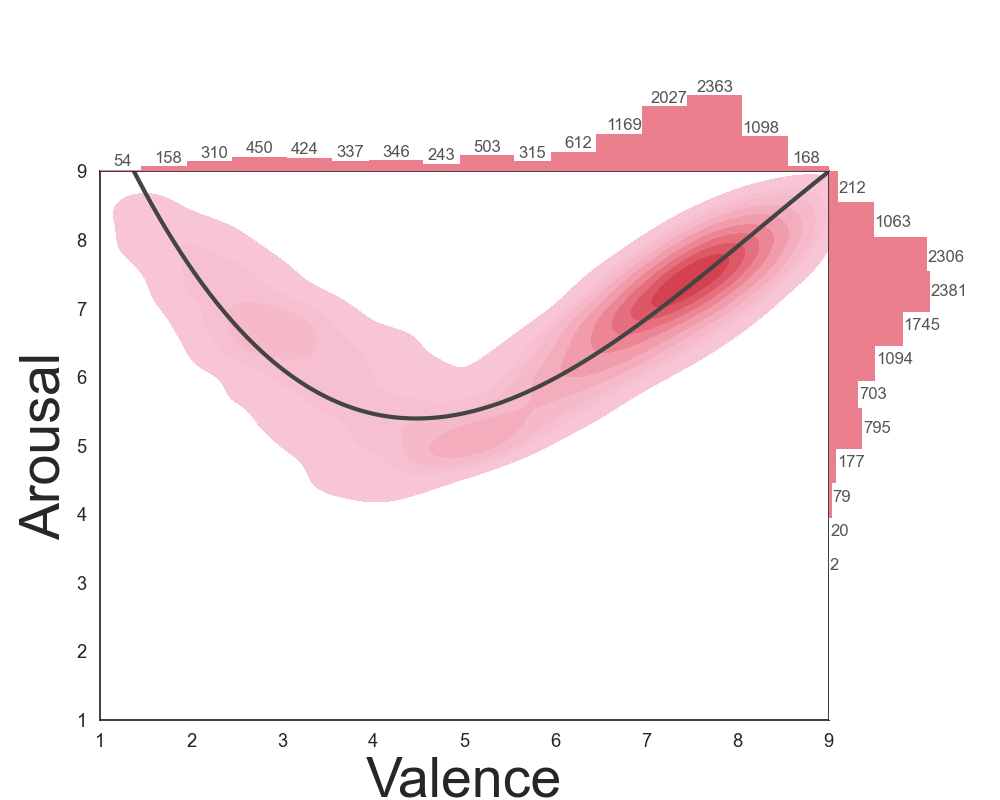}
        \caption{eng-rest}
    \end{subfigure}
    \begin{subfigure}[b]{0.19\textwidth}
        \centering
        \includegraphics[width=\textwidth]{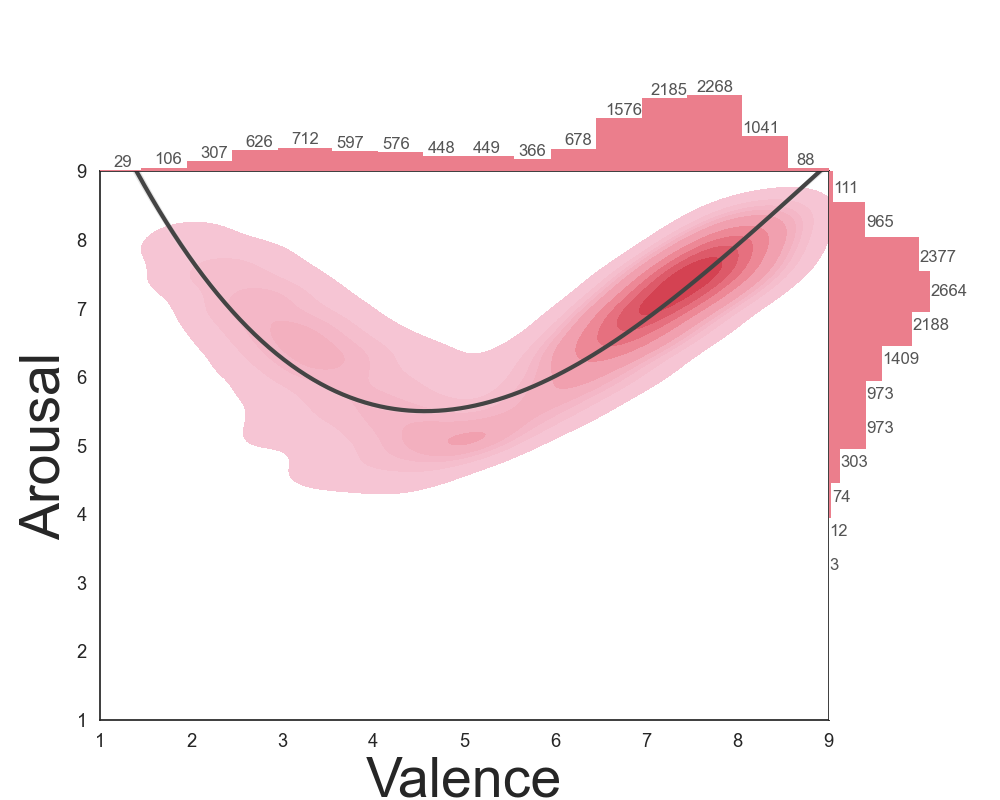}
        \caption{eng-lap}
    \end{subfigure}
    \begin{subfigure}[b]{0.19\textwidth}
        \centering
        \includegraphics[width=\textwidth]{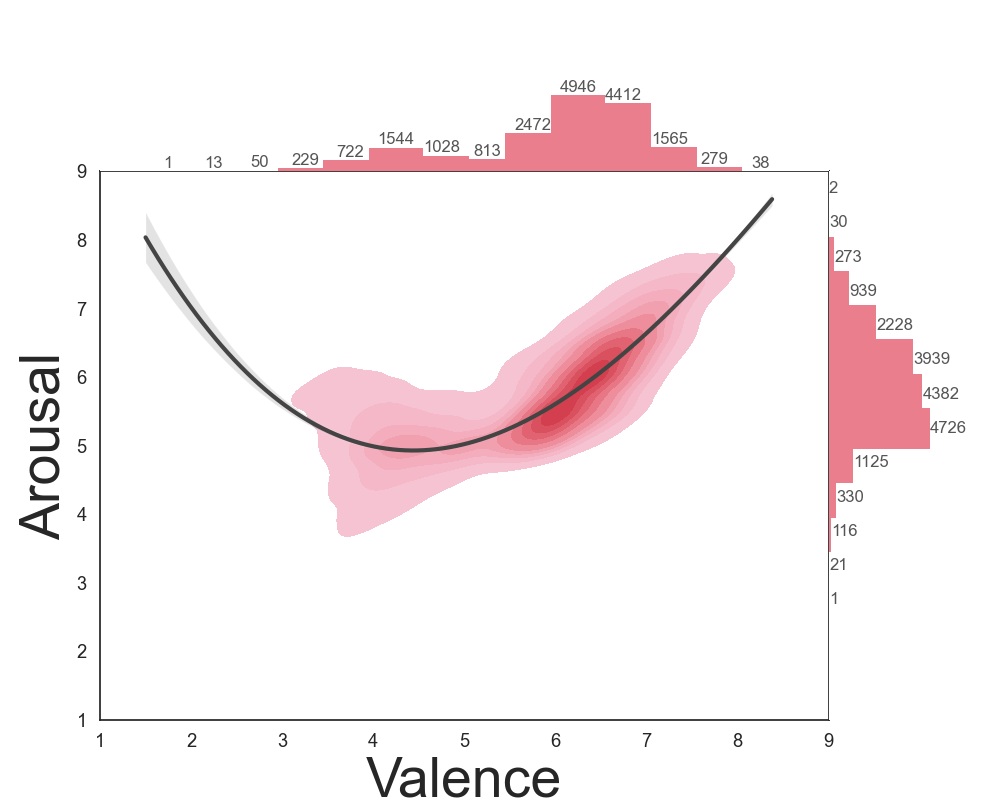}
        \caption{zho-rest}
    \end{subfigure}
    \begin{subfigure}[b]{0.19\textwidth}
        \centering
        \includegraphics[width=\textwidth]{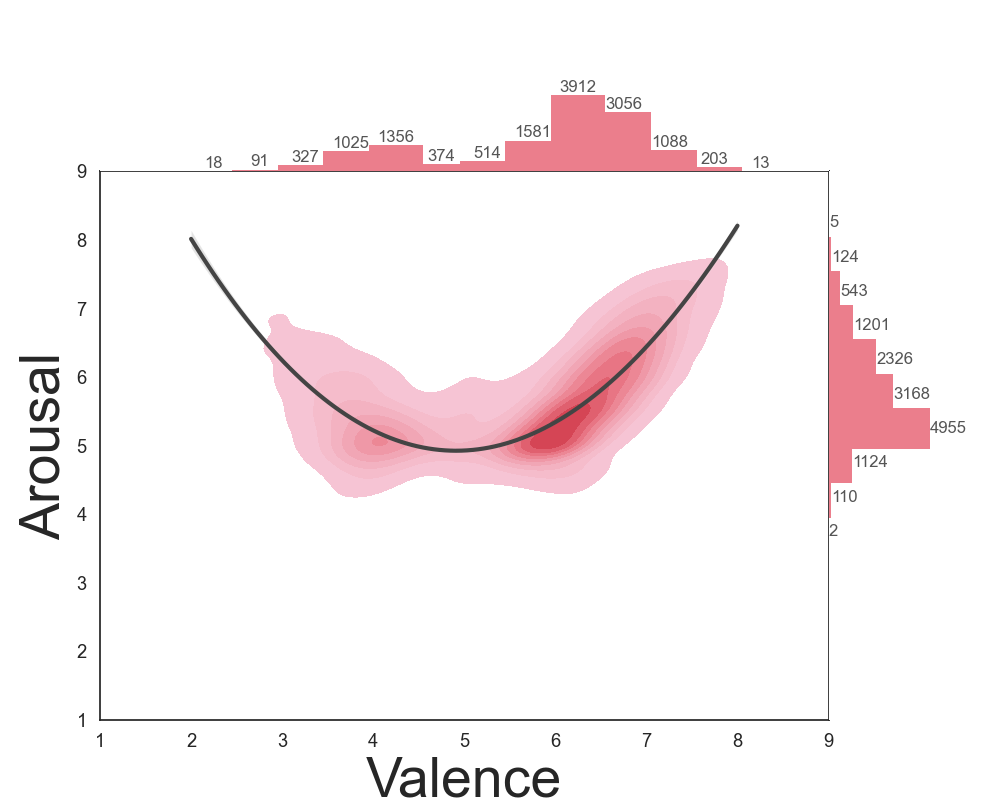}
        \caption{zho-lap}
    \end{subfigure}
    \begin{subfigure}[b]{0.19\textwidth}
        \centering
        \includegraphics[width=\textwidth]{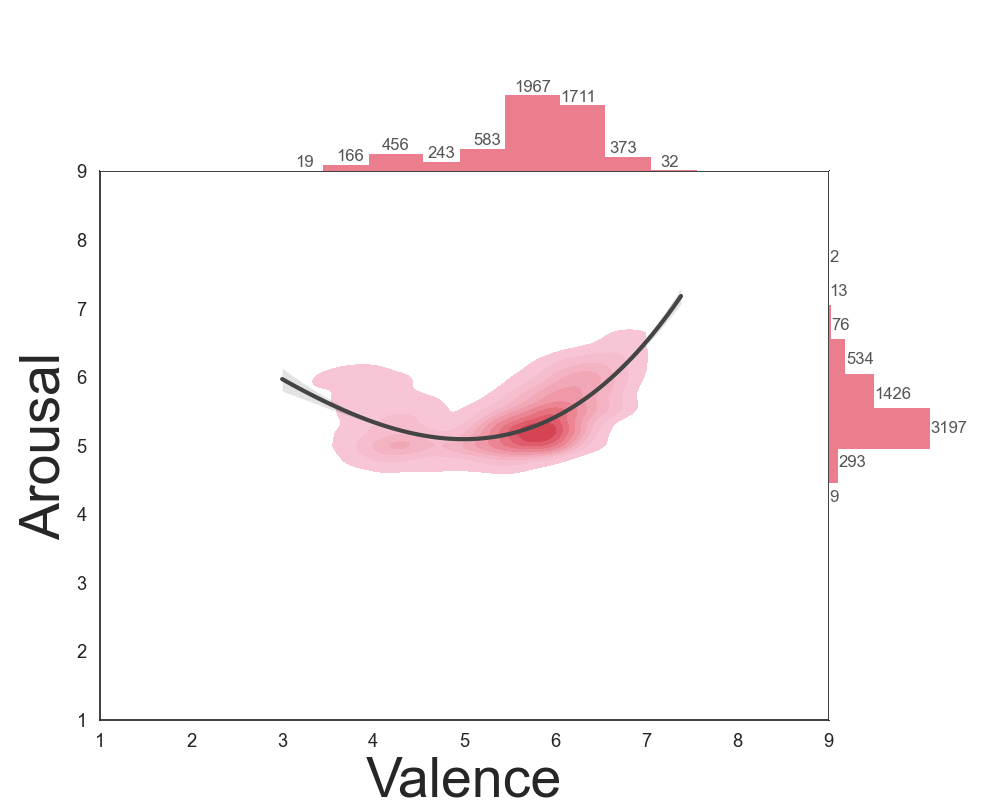}
        \caption{zho-fin}
    \end{subfigure}

    \vspace{0.8em}

    \begin{subfigure}[b]{0.19\textwidth}
        \centering
        \includegraphics[width=\textwidth]{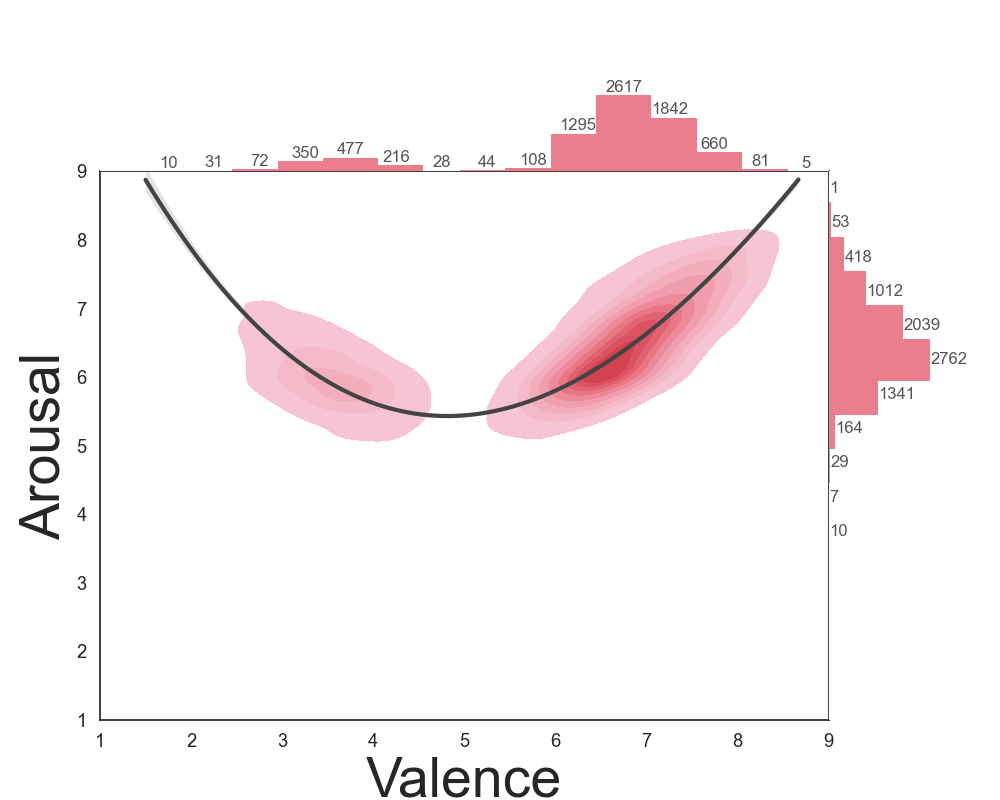}
        \caption{jpn-hot}
    \end{subfigure}
    \begin{subfigure}[b]{0.19\textwidth}
        \centering
        \includegraphics[width=\textwidth]{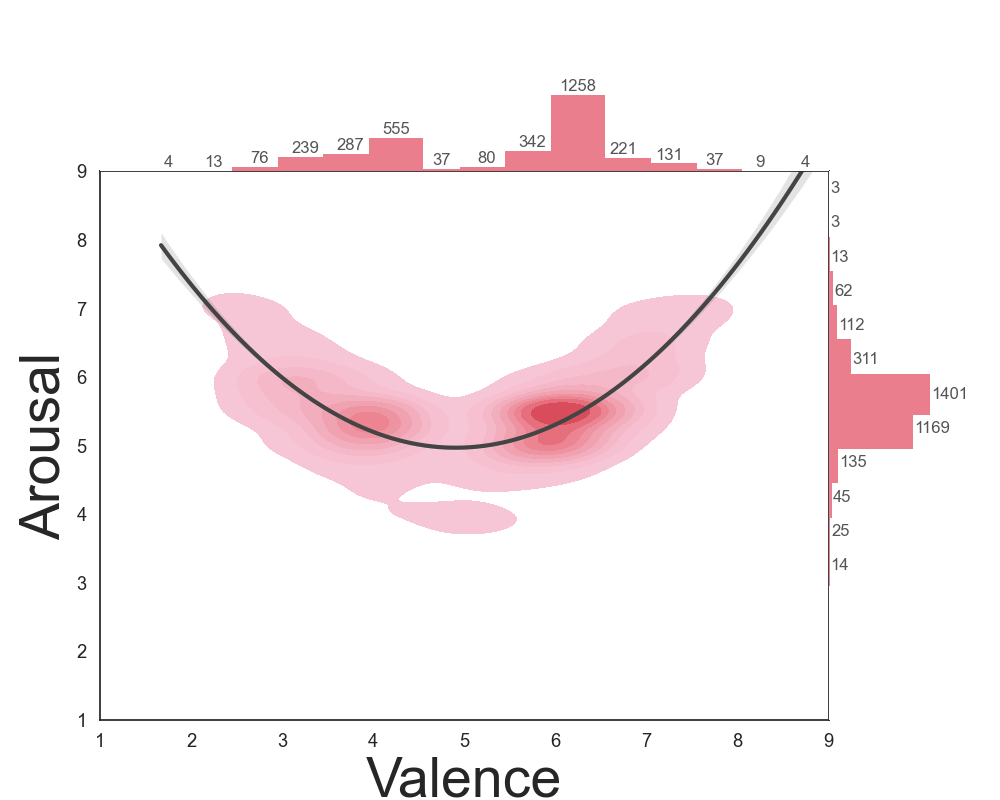}
        \caption{jpn-fin}
    \end{subfigure}
    \begin{subfigure}[b]{0.19\textwidth}
        \centering
        \includegraphics[width=\textwidth]{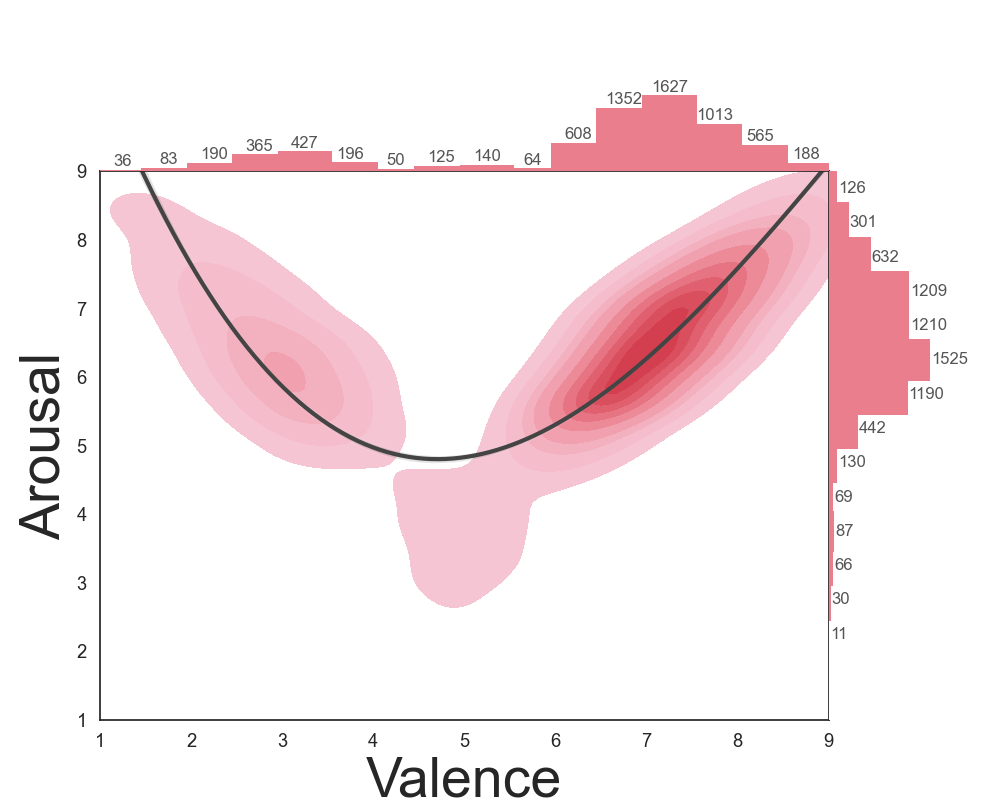}
        \caption{rus-rest}
    \end{subfigure}
    \begin{subfigure}[b]{0.19\textwidth}
        \centering
        \includegraphics[width=\textwidth]{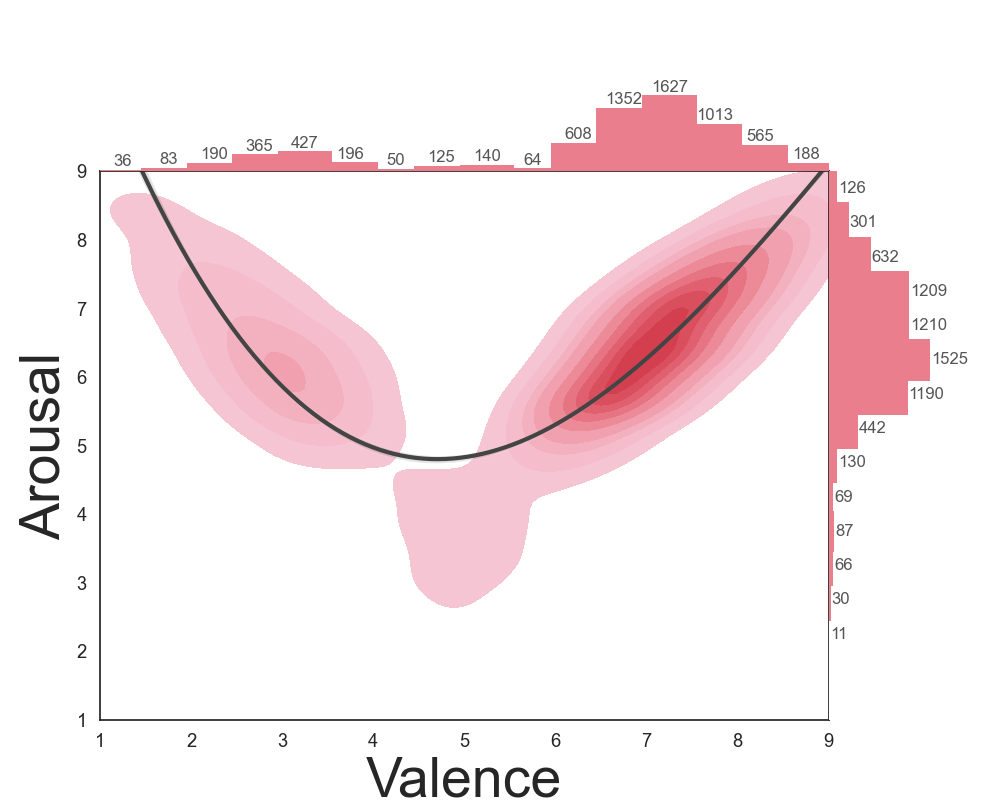}
        \caption{tat-rest}
    \end{subfigure}
    \begin{subfigure}[b]{0.19\textwidth}
        \centering
        \includegraphics[width=\textwidth]{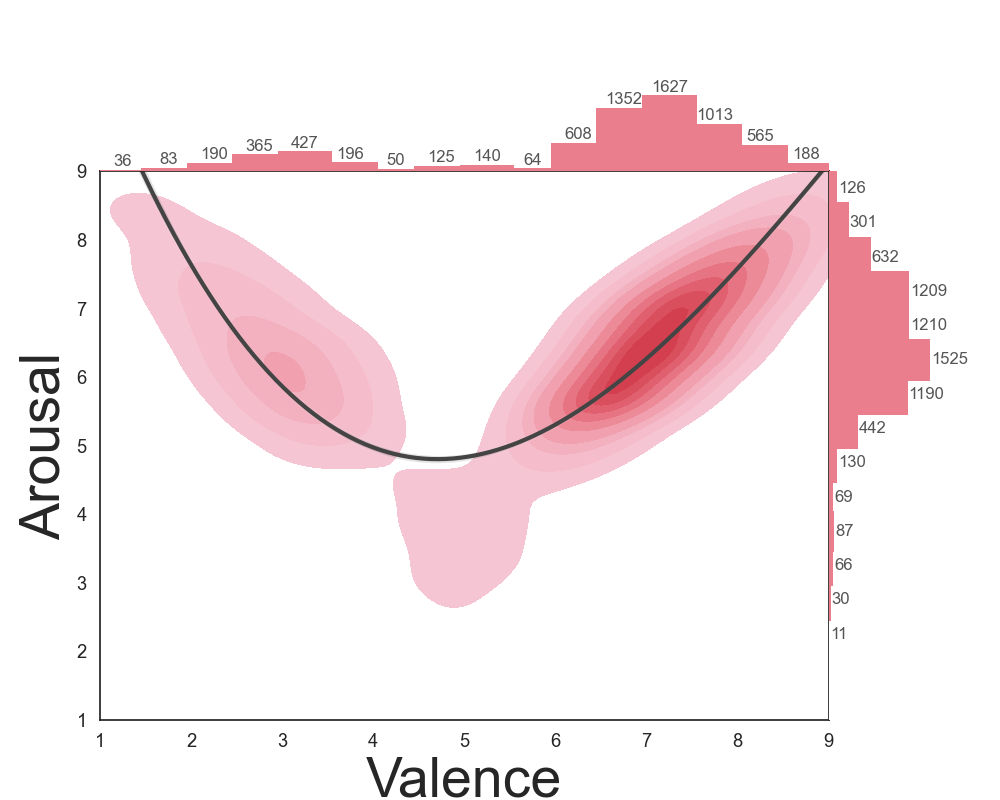}
        \caption{ukr-rest}
    \end{subfigure}

    \vspace{0.8em}

    \caption{\textbf{Joint Valence-Arousal (VA) distributions} across various languages in the DimABSA dataset.}
    \label{fig:va_scatter}
\end{figure*}

\subsection{Evaluating Annotation Quality}
\Cref{tab:dataset_stats_aclstyle} report tuple-level annotation agreement using the F1 score, following prior work~\cite{Chebolu2024, Wu2025}. The F1 score is computed between two annotators by treating one as the predicted output and the other as the gold standard. For VA agreement, we report Root Mean Square Error (RMSE) separately for valence and arousal. RMSE is calculated by comparing each annotator's rating against the mean of the five annotators, and the final agreement score is the average RMSE across all annotators. %

Tuple-level F1 scores show that annotation agreement tends to decrease with increasing tuple complexity. Agreement is highest for single aspect terms, while lower for finance-domain datasets than for customer reviews, likely reflecting domain-specific ambiguity. Agreement on \texttt{(A, O)} pairs remains relatively consistent across datasets. In contrast, agreement on \texttt{(A, C, O)} triplets varies more widely, likely due to the larger and more complex aspect category sets.

For VA annotation, the Japanese and Chinese datasets yield lower RMSE, especially for arousal, likely due to the more formal nature of the financial texts and potential cultural differences in emotional expression. Additionally, annotators consistently exhibit lower agreement for arousal than for valence, confirming prior findings that arousal is harder to annotate reliably~\cite{Buechel2017, mohammad2018obtaining, Lee2022}.

\section{Dataset Analysis}

We analyze the distribution of aspect categories and VA scores across the datasets.

\paragraph{Long-tailed Aspect Category Distribution.} As shown in \Cref{fig:category_scatter}, the distributions across domains are highly imbalanced and exhibit a long-tailed pattern, consistent with findings from previous ABSA tasks \cite{pontiki-etal-2015-semeval, pontiki2016semeval}. 
For the restaurant domain, the 18 categories are highly concentrated: the top five categories account for 97.89\% of all aspect instances, while the remaining 13 categories contribute only 2.11\%.
In contrast, the laptop domain includes 148 categories, with the top five and ten categories covering  37.20\% and 54.18\% of instances, respectively. Similarly, in the hotel domain, the top five categories account for 51\% of the 47 categories (see Appendix~\ref{longtail})

\paragraph{U-shaped VA Distribution.} In \Cref{fig:va_scatter}, the red shaded regions illustrate the joint VA distribution of aspect instances, with darker colors indicating a higher instance density. The black curve represents the relationship between valence and arousal. Overall, all datasets exhibit a broadly U-like VA pattern, indicating that arousal tends to be lowest around neutral valence and increases toward both negative and positive extremes.  

Beyond this shared pattern,we observe variations  across languages and domains. At the language level, Chinese and Japanese exhibit more compact VA distributions with lower dispersion, whereas other languages show greater coverage of both VA extremes.
At the domain level, the finance domain shows a more constrained arousal distribution, with fewer instances at the arousal extremes. This pattern may be attributed to the formal nature of the source data, which consists primarily of financial reports.

\section{Evaluation}

\subsection{Setups}
\paragraph{Tasks.} We evaluate the datasets on three subtasks: DimASR, DimASTE, and DimASQP. The data split sizes are detailed in \Cref{tab:dataset_stats_aclstyle}.

\paragraph{Models.} We evaluate various LLMs under the following two settings. 

\textbf{Zero-/few-shot learning.} We employ two closed-source LLMs, GPT-5 mini \cite{OpenAIGPT-52025} and Kimi K2 Thinking \cite{MoonshotAI2025}. In few-shot settings, in-context examples are selected from the first \textit{$k$} samples in the training set. 
We access both models via API to leverage their built-in reasoning capabilities. The prompts are provided in Appendix~\ref{sec:fewshot_prompt}.
  
\textbf{Supervised fine-tuning.} We fine-tune four publicly available LLMs: Qwen3 14B\footnote{\url{https://huggingface.co/Qwen/Qwen3-14B}}\cite{Alibaba2025}, Ministral-3 14B\footnote{\url{https://huggingface.co/mistralai/Ministral-3-14B-Reasoning-2512}}\cite{MistralAI2025}, Llama-3.3 70B\footnote{\url{https://huggingface.co/meta-llama/Llama-3.3-70B-Instruct}}\cite{Meta2024}, and GPT-OSS 120B\footnote{\url{https://huggingface.co/openai/gpt-oss-120b}}\cite{OpenAI2025}. All models are fine-tuned using 4-bit QLoRA~\cite{Dettmers2023} for parameter-efficient adaptation. The prompts are provided in Appendix~\ref{sec:sft_prompt}.

\paragraph{Implementation Details.} All experiments are implemented using the PyTorch-based implementations of LLMs provided by the Hugging Face Transformers library. We adopt the AdamW optimizer with a linear learning rate scheduler. The learning rate is fixed at 2e-5. The batch size is set to 4 for LLMs. All models are fine-tuned for 5 training epochs. Experiments are conducted on NVIDIA H200 GPUs.

\subsection{Metrics}
\paragraph{Subtask 1.}  DimASR is evaluated by measuring the prediction error in the VA space using RMSE, defined as
\vspace{-0.25cm}
\begin{equation}
\small
\mathrm{RMSE}_\mathrm{{VA}} =
\sqrt{
\frac{1}{N}
\sum_{i=1}^{N}
\left(V_{p}^{(i)}-V_{g}^{(i)}\right)^2+
\left(A_{p}^{(i)}-A_{g}^{(i)}\right)^2
}
\end{equation}

 where $N$ is the total number of instances; $V_p^{(i)}$ and $A_p^{(i)}$ denote the predicted valence and arousal values for an instance; and $V_g^{(i)}$ and $A_g^{(i)}$ denote the corresponding gold values.

\paragraph{Subtasks 2 \& 3.} DimASTE and DimASQP are evaluated using the \textit{continuous F1 (cF1)} score, which unifies categorical and continuous evaluation. Following the standard F1, a predicted tuple is counted as a true positive (TP) only if all its categorical elements exactly match the gold annotation. This categorical TP is then extended as a \textit{continuous true positive (cTP)} by incorporating the VA prediction error. Formally, let $P$ denote the set of predicted triplets \texttt{(A, O, VA)} or quadruplets \mbox{\texttt{(A, C, O, VA)}}. For any prediction $t \in P$, its cTP is defined as
\begin{equation}
\small
\mathrm{cTP}^{(t)} =
\begin{cases} 
1 - \mathrm{dist}(\mathrm{VA}_p^{(t)}, \mathrm{VA}_g^{(t)}), & t \in P_\mathrm{cat} \\
0, & \text{otherwise}
\end{cases}
\end{equation}

 where $P_{cat} \subseteq P$ denotes the set of predictions in which all categorical elements, \texttt{(A, O)} for a triplet or \texttt{(A, C, O)} for a quadruplet, exactly match the gold annotation for the same sentence. The distance function is defined as
\begin{equation}
\small
\mathrm{dist}(\mathrm{VA}_{p}, \mathrm{VA}_{g}) =
\frac{
\sqrt{(V_p - V_g)^2 + (A_p - A_g)^2}
}{
\mathrm{D}_{\max}
}
\end{equation}

 where $\operatorname{dist}(\cdot)$ denotes the normalized Euclidean distance between the predicted $\mathrm{VA}_p = (V_p, A_p)$ and gold $\mathrm{VA}_{g}=(V_g,A_g)$ in the VA space, and $\mathrm{D}_{\max} = \sqrt{8^2+8^2} = \sqrt{128}$ is the maximum possible Euclidean distance in the VA space on the [1, 9] scale, ensuring that $\operatorname{dist} \in [0, 1]$.

Building on per-prediction $\mathrm{cTP}^{(t)}$, \textit{cRecall} and \textit{cPrecision} are defined as the total cTP divided by the number of gold and predicted triplets/quadruplets, respectively. The cF1 is computed as their harmonic mean. An illustrative example is given in 
Appendix~\ref{sec:example_calculating_cF1}

\begin{table*}[t]
\centering
\small
\setlength{\tabcolsep}{2.2pt}
\renewcommand{\arraystretch}{1.0}

\newcommand{\bestASR}[1]{\cellcolor{blue!25}\textbf{#1}}
\newcommand{\bestASTE}[1]{\cellcolor{red!25}\textbf{#1}}
\newcommand{\bestASQP}[1]{\cellcolor{yellow!25}\textbf{#1}}
\begin{tabular}{ll cc|cc |cccc}
\toprule
&& \multicolumn{2}{c|}{Zero-Shot Learning} 
& \multicolumn{2}{c|}{One-Shot Learning} 
& \multicolumn{4}{c}{Supervised Fine-Tuning}\\
\cmidrule(lr){3-4} \cmidrule(lr){5-6} \cmidrule(lr){7-10}
Subtask&Dataset & \multicolumn{1}{p{1.5cm}|}{\centering GPT-5 \\mini \\ } & \multicolumn{1}{p{1.5cm}|}{\centering Kimi K2 \\ Thinking} & \multicolumn{1}{p{1.5cm}|}{\centering GPT-5 \\mini \\ } & \multicolumn{1}{p{1.5cm}|}{\centering Kimi K2 \\ Thinking} & \multicolumn{1}{p{1.5cm}}{\centering Qwen3 \\ (14B)} & \multicolumn{1}{p{1.5cm}}{\centering Ministral-3 \\ (14B)} & \multicolumn{1}{p{1.5cm}}{\centering Llama-3.3 \\ (70B)} & \multicolumn{1}{p{1.5cm}}{\centering GPT-OSS \\ (120B)} \\ \midrule
\multirow{10}{*}{DimASR} & eng-rest & 2.9490 & \bestASR{2.3432} & 2.3926 & \bestASR{2.1461} & 2.6427 & 2.6316 & 2.5244 & \bestASR{1.4605} \\
&eng-lap & 3.2115 & \bestASR{2.6546} & 2.5637 & \bestASR{2.1893} & 2.8089 & 2.6258 & 2.7354 & \bestASR{1.5269} \\
&jpn-hot & 3.1406 & \bestASR{2.3294} & 2.1607 & \bestASR{1.7553} & 2.2906 & 2.2999 & 2.6255 & \bestASR{0.7188}\\
&jpn-fin & 2.6760 & \bestASR{2.3379} & 1.9243 & \bestASR{1.6396} & 1.8964 & 2.0700 & 2.4191 & \bestASR{1.0188}\\
&rus-rest & 2.5447 & \bestASR{2.0630} & 2.0390 & \bestASR{1.7768} & 2.1528 & 2.3617 & 2.5089 & \bestASR{1.4775}\\
&tat-rest & 2.6645 & \bestASR{2.3636} & 2.2308 & \bestASR{1.9380} & 2.6367 & 3.0463 & 2.9165 & \bestASR{1.7153}\\
&ukr-rest & 2.5628 & \bestASR{2.0782} & 2.0438 & \bestASR{1.7805} & 2.2121 & 2.4592 & 2.5709 & \bestASR{1.5166}\\
&zho-rest & 2.7125 & \bestASR{2.2623} & 2.2467 & \bestASR{1.8959} & 2.0073 & 1.9373 & 2.4463 & \bestASR{1.0349}\\
&zho-lap & 2.4790 & \bestASR{2.0426} & 1.9380 & \bestASR{1.6440} & 1.7706 & 1.8267 & 2.3633 &\bestASR{0.8032} \\
&zho-fin & \bestASR{2.6547} & 2.9662 & 2.0094 & \bestASR{1.9652} & 1.4707 & 1.8900 & 2.5632 & \bestASR{0.6511}\\
\cmidrule(lr){1-10}
&AVG & 2.7595 & \bestASR{2.3441} & 2.1549 & \bestASR{1.8731} & 2.1889 & 2.3149 & 2.5674 & \bestASR{1.1924} \\
\cmidrule(lr){1-10}\morecmidrules\cmidrule(lr){1-10}
\multirow{8}{*}{DimASTE} & eng-rest & 0.4993 & \bestASTE{0.5101} & \bestASTE{0.5034} & 0.4920 & 0.4483 & 0.2930 & 0.5418 & \bestASTE{0.5442} \\
&eng-lap & 0.4491 & \bestASTE{0.4519} & \bestASTE{0.4874} & 0.4424 & 0.3827 & 0.2736 & \bestASTE{0.4664} & 0.4515 \\
&jpn-hot & 0.1727 & \bestASTE{0.3148} & 0.2487 & \bestASTE{0.3464} & 0.1622 & 0.1458 & 0.4694 & \bestASTE{0.5397} \\
&rus-rest & 0.4016 & \bestASTE{0.4031} & 0.3730 & \bestASTE{0.4242} & 0.3341 & 0.1774 & \bestASTE{0.4590} & 0.4262 \\
&tat-rest & 0.3419 & \bestASTE{0.3503} & 0.3159 & \bestASTE{0.3577} & 0.2020 & 0.1154 & \bestASTE{0.4101} & 0.3578 \\
&ukr-rest & 0.4014 & \bestASTE{0.4082} & 0.3326 & \bestASTE{0.4220} & 0.3099 & 0.1595 & \bestASTE{0.4517} & 0.4250 \\
&zho-rest & 0.3190 & \bestASTE{0.3728} & 0.2425 & \bestASTE{0.3529} & 0.2509 & 0.1446 & \bestASTE{0.4789} & 0.4759 \\
&zho-lap & \bestASTE{0.2372} & 0.2227 & \bestASTE{0.2815} & 0.2494 & 0.2099 & 0.1182 & 0.4344 & \bestASTE{0.4366} \\
\cmidrule(lr){1-10}
&AVG & 0.3528 & \bestASTE{0.3792} & 0.3481 & \bestASTE{0.3859} & 0.2875 & 0.1784 &  \bestASTE{0.4640} & 0.4571 \\
\cmidrule(lr){1-10}\morecmidrules\cmidrule(lr){1-10}
\multirow{8}{*}{DimASQP} & eng-rest & \bestASQP{0.4036} & 0.3740 & \bestASQP{0.3816} & 0.3746 & 0.2673 & 0.2058 & \bestASQP{0.5048} & 0.5013 \\
&eng-lap & 0.2304 & \bestASQP{0.2805} & \bestASQP{0.2842} & 0.2795 & 0.1529 & 0.1293 & \bestASQP{0.2483} & 0.2411 \\
&jpn-hot & 0.0907 & \bestASQP{0.1309} & 0.1542 & \bestASQP{0.1943} & 0.0400 & 0.0311 & 0.3577 & \bestASQP{0.4151} \\
&rus-rest & 0.2508 & \bestASQP{0.2656} & 0.2505 & \bestASQP{0.2963} & 0.1682 & 0.1000 & \bestASQP{0.4118} & 0.3683\\
&tat-rest & 0.1974 & \bestASQP{0.2310} & 0.2025 & \bestASQP{0.2380} & 0.0954 & 0.0611 & \bestASQP{0.3702} & 0.3094 \\
&ukr-rest & 0.2465 & \bestASQP{0.2974} & 0.2180 & \bestASQP{0.2971} & 0.1641 & 0.0975 & \bestASQP{0.4070} & 0.3663 \\
&zho-rest & 0.2481 & \bestASQP{0.2975} & 0.1891 & \bestASQP{0.2859} & 0.1605 & 0.0934 & \bestASQP{0.4391} & 0.4249 \\
&zho-lap & 0.1356 & \bestASQP{0.1569} & \bestASQP{0.1921} & 0.1900 & 0.1124 & 0.0728 & 0.3506 & \bestASQP{0.3551} \\
\cmidrule(lr){1-10}
&AVG & 0.2254 & \bestASQP{0.2542} & 0.2340 & \bestASQP{0.2695} & 0.1451 & 0.0989 & \bestASQP{0.3862} & 0.3727 \\
\bottomrule
\end{tabular}
\caption{\textbf{Comparison of LLM-based approaches}. RMSE is used for the DimASR subtask, while cF1 is used for the DimASTE and DimASQP subtasks. Bold indicates the best result within each setting for each dataset. cPrecision and cRecall results for the DimASTE and DimASQP subtasks are shown in Appendix~\ref{DimASTE-ASQPResults}.}
\label{tab:LLMresults}
\end{table*}

\subsection{Results}
\Cref{tab:LLMresults} summarizes the results of various LLMs across all subtasks. Closed-source models serve as data-efficient baselines, with Kimi-K2-Thinking achieving higher average performance than \mbox{GPT-5} mini under both zero- and one-shot settings. Fine-tuned large-scale LLMs further improve performance on average over their closed-source counterparts. Overall, the results suggest that all DimABSA subtasks remain challenging.

\textbf{DimASR.} This subtask presents baseline results for sentiment regression. Closed-source models provide strong prompting baselines. The fine-tuned 14B and 70B models do not consistently outperform these baselines. In contrast, fine-tuning the 120B model yields a substantial and consistent improvement across languages, indicating that supervised adaptation can enhance the model's ability to capture nuanced affective information in multilingual settings.

\begin{figure*}[ht]
    \centering

    \begin{subfigure}[b]{0.32\textwidth}
        \centering
        \includegraphics[width=\textwidth]{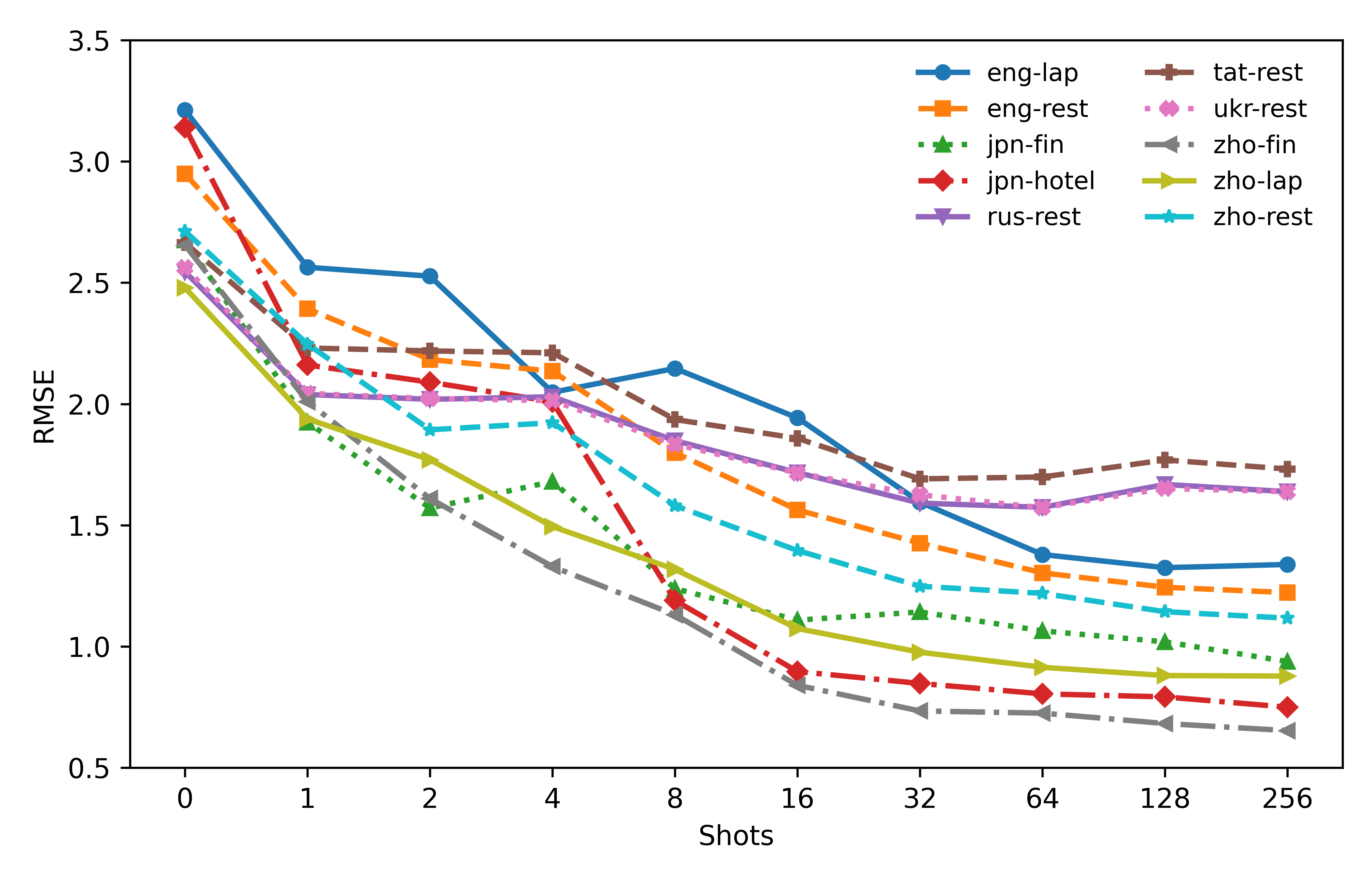}
        \caption{DimASR}
        \label{fig:DimASR_fewshot}
    \end{subfigure}
    \vspace{0.6em}
    \begin{subfigure}[b]{0.32\textwidth}
        \centering
        \includegraphics[width=\textwidth]{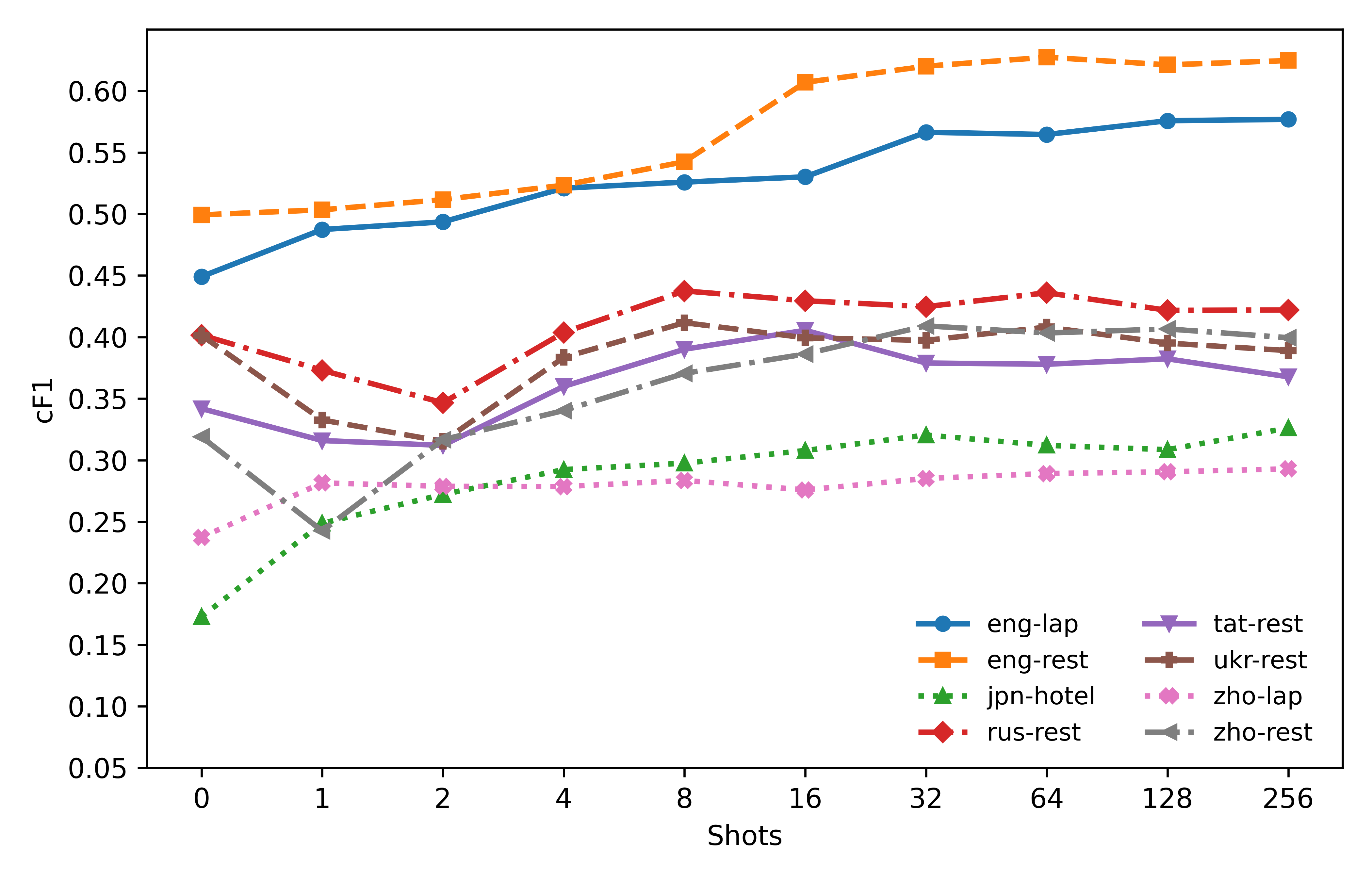}
        \caption{DimASTE}
        \label{fig:DimASTE_fewshot}
    \end{subfigure}
    \vspace{0.6em}
    \begin{subfigure}[b]{0.32\textwidth}
        \centering
        \includegraphics[width=\textwidth]{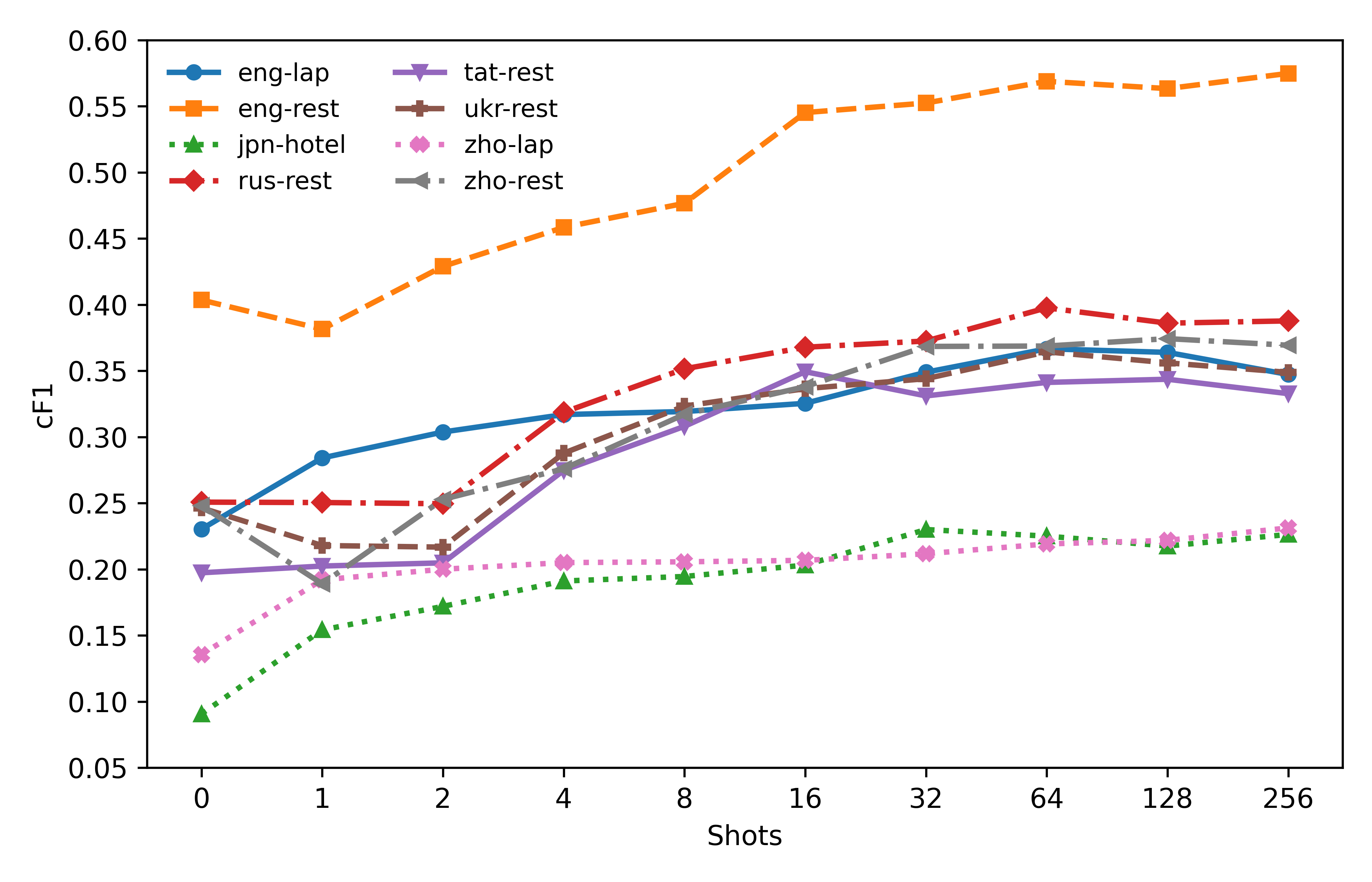}
        \caption{DimASQP}
        \label{fig:DimASQP_fewshot}
    \end{subfigure}
\caption{\textbf{Few-shot Performance of GPT-5 mini.} Results are summarized across all subtasks and datasets. See Appendix~\ref{Few-shotResults} for corresponding numerical scores.}
\label{fig:few_shot}
\vspace{10pt}
\end{figure*}

\setlength{\floatsep}{30pt}
\begin{figure*}[ht]
    \centering
    \begin{subfigure}[b]{0.98\textwidth}
        \centering
        \includegraphics[width=\textwidth]{figures/va_scatter_few-shot/dimasr_eng_restaurant_va_scatter.png}
    \end{subfigure}

    \vspace{0.6em}
    \caption{\textbf{Gold and predicted VA distributions for DimASR.} Results represent the English–Restaurant test set; see Appendix~\ref{Few-shotVAscatter} for other datasets.}
    \label{fig:DimASR_few-shot_va_scatter_eng-rest}
\end{figure*}

\textbf{DimASTE.} This subtask presents baseline results for sentiment regression coupled with structured \mbox{\texttt{(A, O)}} extraction, revealing language-specific differences in LLM performance. Both closed-source models perform best on English, followed by Russian and Ukrainian, with a noticeable gap to other languages, even high-resource ones such as Chinese and Japanese. This suggests that the structural patterns required for extraction in English and Slavic languages are more effectively captured during LLM pretraining.

After fine-tuning, only the 70B and 120B models show consistent improvements, with particularly substantial gains in previously underperforming Chinese and Japanese. This indicates that supervised adaptation at sufficient scale helps compensate for underrepresented structural patterns during pretraining. Notably, although Tatar also benefits from fine-tuning, it remains the lowest-performing language, suggesting that LLMs still struggle with low-resource languages whose structural properties are difficult to capture.

\textbf{DimASQP.} This subtask highlights the impact of introducing aspect categories on extraction performance, revealing degradation across all models, with the fine-tuned 70B and 120B models showing smaller drops. At the domain level, the laptop domain shows a significant decrease, likely due to its larger and more diverse category set, whereas the restaurant domain drops less sharply, reflecting its smaller, more concentrated category set. Within the restaurant domain, English achieves the best performance across languages, while Tatar remains the lowest-performing language, a trend consistent with the DimASTE results.

\subsection{Analysis}
\textbf{Effect of few-shot prompting on performance.}
As shown in \Cref{fig:few_shot}, DimASR behaves differently from DimASTE and DimASQP in the early few-shot setting. The one-shot performance improvement for DimASR indicates that even a single example provides a crucial reference for the continuous VA scale, allowing the model to calibrate its numerical output. In contrast, DimASTE and DimASQP show unstable early performance, with limited examples yielding marginal gains due to the complexity of structural induction. Across subtasks, performance generally reaches a plateau around 32 shots. Results up to 256 shots show that few-shot prompting underperforms fine-tuned benchmarks (GPT-OSS 120B for DimASR and Llama-3.3 70B for DimASTE and DimASQP) on all datasets except English. This suggests that fine-tuning remains a strong alternative to prompting.  

\textbf{Effect of few-shot prompting on VA distributions.}
As shown in \Cref{fig:DimASR_few-shot_va_scatter_eng-rest}, zero shot prompting produces a random grid-like VA distribution across languages, likely due to its token-based generation without reference calibration. In contrast, the one-shot setting immediately begins to align with the gold distribution, indicating initial calibration of the VA scale. As more in-context examples are provided (up to 64 and 256 shots), the predicted distribution increasingly resemble the gold standard. However, although higher shot counts yield closer visual alignment, performance does not necessarily continue to improve once it reaches saturation.

\textbf{Categorical version of the DimABSA dataset.}\footnote{The categorical version of the dataset is available at: \url{https://github.com/DimABSA/DimABSA2026/tree/main/task-dataset/track_a/categorical}.}
To align with traditional ABSA, we partition the DimABSA datasets into positive (V>5.5), neutral (4.5<=V<=5.5), and negative (V<4.5) subsets, and conduct polarity-based evaluation using the same LLMs (see Appendix ~\ref{Catgorical Results}). The results show that the fine-grained DimABSA datasets are more challenging than their categorical counterparts, while this conversion also provides a new multilingual ABSA dataset in the standard categorical setting.

\section{Related Work}
\textbf{Dimensional sentiment resources.} Previous studies on dimensional sentiment analysis have developed resources with single or combined dimensions across lexical, phrasal, and sentential granularities. 
Sentiment lexicons assign affective scores to individual words, such as \mbox{SentiWordNet} \cite{Baccianella2010}, SO-CAL \cite{Taboada2011}, SentiStrength \cite{Thelwall2012}, and NRC-VAD \cite{mohammad2018obtaining, mohammad2025nrcvad}. 
Phrase-level datasets formulate sentiment composition through modifiers, including SemEval-2015 Task 10 \cite{Rosenthal2015} and SemEval-2016 Task 7 \cite{Kiritchenko2016}. 
At the sentence level, affective scores are provided for texts of varying lengths \cite{preotiuc-pietro-etal-2016-modelling, Buechel2017, mohammad-bravo-marquez-2017-emotion, Mohammad2018semeval, muhammad-etal-2025-brighter}. The Stanford Sentiment Treebank \cite{Socher2013} and Chinese EmoBank \cite{Lee2022} provide cross-granularity resources, bridging phrase- and sentence-level representations and covering all three granularities.

\textbf{Traditional ABSA datasets.} Most existing ABSA datasets are English-centric and use categorical polarity labels. SemEval-2014 \cite{pontiki-etal-2014-semeval} initiated ABSA for English restaurant and laptop reviews, followed by extensions to more subtasks and languages \cite{pontiki-etal-2015-semeval, pontiki2016semeval}. 
Subsequent datasets progressively enriched the annotation schema, introducing triplets of aspect, opinion, and polarity \cite{Xu2020, Peng2020}, quadruples by adding an aspect category \cite{Zhang2021, Cai2021}, and broader domain coverage \cite{Chebolu2024}. 
M-ABSA \cite{Wu2025} further introduced a multilingual parallel benchmark via automatic translation.

\textbf{Dimensional ABSA datasets.} Existing work is limited to a single-language, single-domain Chinese restaurant dataset from SIGHAN-2024 \cite{Lee2024}. Our DimABSA datasets are the first multilingual and multi-domain resource for dimensional ABSA.

\section{Conclusions}

We present \datasetname, the first multilingual dimensional ABSA resource with VA annotations. It comprises 10 datasets spanning six high- and low-resource languages across four domains.
To move beyond categorical ABSA, we introduce three subtasks ranging from pure VA regression to joint extraction-regression tasks, and formulate a unified metric for categorical and continuous evaluation. 
Experiments with various LLMs reveal two key limitations: the constraints of token-based VA prediction in regression and persistent difficulties with hybrid extraction–regression tasks, especially for low-resource languages. These findings underscore critical challenges and opportunities for advancing multilingual dimensional ABSA.

\section*{Limitations}
Although DimABSA is multilingual, interpretations of valence and arousal can vary across cultures, thereby affecting cross-lingual comparability.
We mitigate this by using five native-speaker annotators per language and sample, consistent 1–9 VA scales, and shared guidelines; nonetheless, results should be interpreted as comparisons across language-community-domain settings. Expanding language coverage and testing measurement invariance are important directions for future work.

Cross-cultural comparison is a limitation of all data. For example, people from one culture may be prone to give annotations close to the centre of the scale whereas people in another culture may be more likely to spread their responses. Also, people from different cultures may use different amounts of emotional language to convey the same intensity of emotion. Therefore, cross-cultural perception is an essential issue for further research.

\section*{Ethical Considerations}

People expresses attitudes, opinions, and sentiment towards entities and their aspects in complex and nuanced ways. Further, there can be considerable person-to-person variation. It should be noted that human annotated labels capture \textbf{perceived} sentiment and attitudes, and that in several cases this may be different from the speaker's true attitudes. Nonetheless, since language is key mechanism to communicate, at an aggregate-level perceived opinions tend to correlate with actual opinions. Thus, even perceived opinions are useful at an aggregate level. However, caution must be employed when using individual inferred opinions to make decisions about individuals, especially high-stakes decisions.  

Like various enabling technologies, ABSA can be abused and misused. Notably, for example, it can be used to identify various likes and dislikes of peoples and use it to manipulate people into behaviour that may not be in their best interest (e.g., purchasing products or availing services that they cannot afford or that are not particularly useful to them). This is especially concerning for vulnerable populations such as children and elderly. We expressly forbid any commercial use of our data.

For a detailed discussion of a large number of ethical considerations associated with automatic sentiment and emotion detection, we refer the reader to \cite{mohammad-2022-ethics-sheet,mohammad-2023-best}.

\section*{Use of AI Assistants} 
We used ChatGPT and Grammarly to correct grammatical errors, enhancing the coherence of the final manuscript. While these tools have augmented our capabilities and contributed to our findings, it\textquoteright s pertinent to note that they have inherent limitations. We have made every effort to use AI in a transparent and responsible manner. Any conclusions drawn are a result of combined human and machine insights. 

\section*{Acknowledgments}
Lung-Hao Lee and Liang-Chih Yu acknowledge the support of National Science and Technology Council, Taiwan, under the grants NSTC 113-2221-E-155-046-MY3 and NSTC 114-2221-E-A49-059-MY3.

Jonas Becker acknowledges the support of the Landeskriminalamt NRW.
Jan Philip Wahle and Terry Ruas acknowledge the support of the Lower Saxony Ministry of Science and Culture, and the VW Foundation.

Shamsuddeen Hassan Muhammad acknowledges the support of Google DeepMind, whose funding made this work possible.

The work of Alexander Panchenko and Ilseyar Alimova was supported by the grant for research centers in the field of AI provided by the Ministry of Economic Development of the R.F. in accordance with the agreement 000000C313925P4F0002 and the agreement with Skoltech №139-10-2025-033. 

Ilseyar Alimova gratefully acknowledges Dina Abdullina for the Tatar data annotation.

\bibliography{custom, DimABSA}

\appendix

\onecolumn
\section*{Appendix}
\label{sec:appendix}
\raggedbottom

\section{Dataset Examples}
\begin{table}[H]
\centering
\small
\renewcommand{\arraystretch}{1.5} 
\begin{tabularx}{\textwidth}{c c p{5cm} X} 
\toprule
\textbf{Dataset} & \textbf{ST} & \textbf{Input} & \textbf{Output} \\ \midrule

\multirow{3}{*}{eng-rest} & ST 1 & The \underline{food} was excellent & 8.00\#8.12 \\ \cmidrule(lr){2-4}
 & ST 2 & Service at the bar was a little slow & (Service, a little slow, 4.10\#4.30) \\ \cmidrule(lr){2-4}
 & ST 3 & Their sodas are usually expired and flat & (sodas, DRINKS\#QUALITY, usually expired, 1.90\#7.20) \par (sodas, DRINKS\#QUALITY, flat, 2.40\#6.80) \\ \hline

\multirow{4}{*}{eng-lap} & ST 1 & Very poor \underline{build quality} & 2.25\#8.12 \\ \cmidrule(lr){2-4}
 & ST 2 & Display is bright and great & (Display, bright, 7.00\#7.50) \par (Display, great, 7.75\#8.25) \\ \cmidrule(lr){2-4}
 & ST 3 & This HP laptop is as good as it sounds on paper & (HP laptop, LAPTOP\#GENERAL, good, 5.75\#5.50) \\ \hline
 
\multirow{7}{*}{jpn-hot} & ST 1 & 強風のせいか建物の \underline{軋む音}がうるさすぎて眠れなかった \par (Because of the strong wind, the creaking sounds of the building were too loud, and I couldn’t sleep.) & 2.38\#7.5 \\ \cmidrule(lr){2-4}
& ST 2 & 作りは古く特別な部分は無いが、立地などなど、総合的には及第點 \par (The building is quite old and has nothing particularly special about it, but its location and other factors make it acceptable overall.) & (綜合的, 及第點, 6.00\#5.38) \par (作り, 古く, 4.17\#4.83)\\ \cmidrule(lr){2-4}
& ST 3 & 駅近なのでとても便利です \par (Because it is close to the station, it is very convenient.)& (駅近, LOCATION\#GENERAL, とても便利, 7.30\#7.20) \\ \hline

jpn-fin & ST 1 & 主な減少は、\underline{投資有価証券}が11億54百万円であります。 \par (The main decrease amounted to 1.154 billion yen in investment securities.) & 3.75\#5.00 \\ \hline

\multirow{3}{*}{rus-res} & ST 1 & \includegraphics[width=\linewidth]{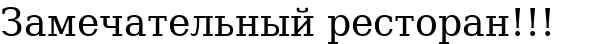} \par (A wonderful \underline{restaurant}!!!) & 9.00\#9.00 \\ \cmidrule(lr){2-4}
& ST 2 & \includegraphics[width=\linewidth]{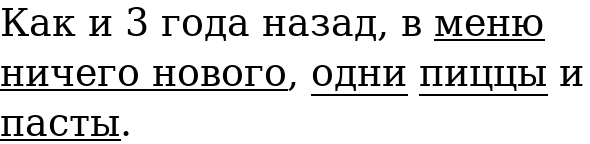} \par (Just like three years ago, there is \underline{nothing new} on the \underline{menu}, \underline{only} \underline{pizzas} and \underline{pastas}.) & \includegraphics[width=\linewidth]{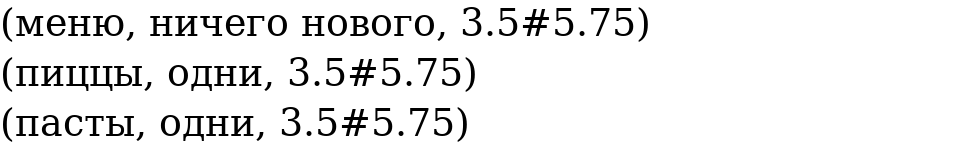}\\ \cmidrule(lr){2-4}
& ST 3 & \includegraphics[width=\linewidth]{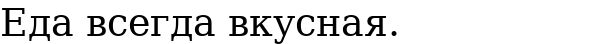} \par (The \underline{food} is \underline{always delicious}.)& \includegraphics[width=\linewidth]{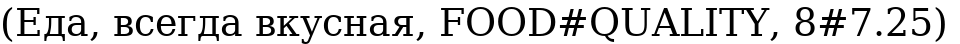} \\ \hline

\multirow{3}{*}{tat-res} & ST 1 & \includegraphics[width=\linewidth]{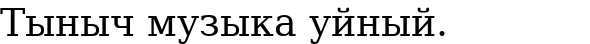} \par (Quiet \underline{music} is playing.) & 5.17\#3.33 \\ \cmidrule(lr){2-4}
& ST 2 & \includegraphics[width=\linewidth]{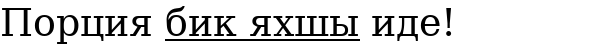} \par (The \underline{portion} was \underline{very good}!) & \includegraphics[width=\linewidth]{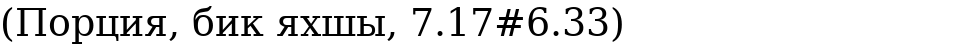}\\ \cmidrule(lr){2-4}
& ST 3 & \includegraphics[width=\linewidth]{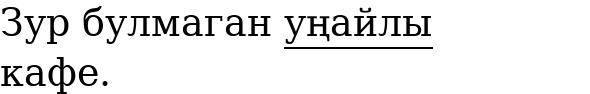} \par (A small, \underline{comfortable} \underline{cafe}.)& \includegraphics[width=\linewidth]{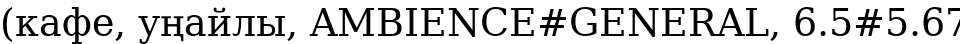}\\
\bottomrule
\end{tabularx}
\end{table}

\begin{table}[H]
\centering
\small
\renewcommand{\arraystretch}{1.5} 
\begin{tabularx}{\textwidth}{c c p{5cm} X} 
\toprule
\textbf{Dataset} & \textbf{ST} & \textbf{Input} & \textbf{Output} \\ \midrule
\multirow{3}{*}{ukr-res} & ST 1 & \includegraphics[width=\linewidth]{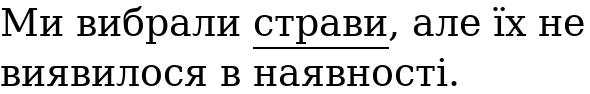}\par (We chose \underline{dishes}, but they were not available.) & 3.00\#6.00 \\ \cmidrule(lr){2-4}
& ST 2 & \includegraphics[width=\linewidth]{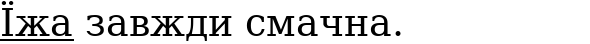} \par (The \underline{food} is \underline{always tasty}.) & \includegraphics[width=\linewidth]{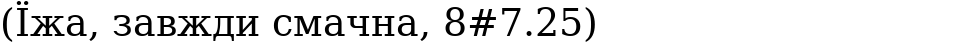}\\ \cmidrule(lr){2-4}
& ST 3 & \includegraphics[width=\linewidth]{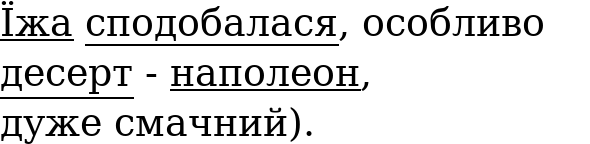} \par (We \underline{liked} the \underline{food}, especially the \underline{dessert} \underline{Napoleon}. It's \underline{very tasty}).)& \includegraphics[width=\linewidth]{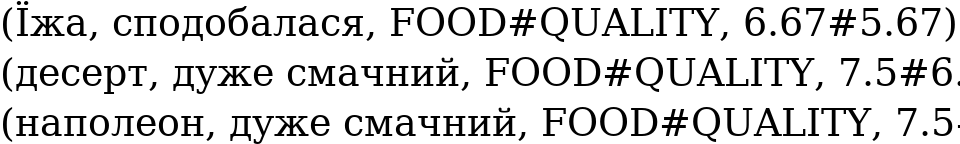}\\ \hline

\multirow{3}{*}{zho-res} & ST 1 & 但是\underline{味道}真的非常普通。\par (But the taste is really very ordinary.) & 4.00\#6.00 \\ \cmidrule(lr){2-4}
& ST 2 & 服務那裡好?根本爛透了。 \par (How was the service there? It was absolutely terrible.) & (服務, 根本爛透, 2.75\#7.12)\\ \cmidrule(lr){2-4}
& ST 3 & 鮮奶凍很濃郁、小湯圓也很好吃，環境乾淨明亮。\par (The fresh milk pudding is very rich, the small glutinous rice balls are also delicious, and the environment is clean and bright.)& (鮮奶凍, FOOD\#QUALITY, 很濃郁, 6.67\#6.17)\par (小湯圓, FOOD\#QUALITY, 很好吃, 7.00\#6.25)\par (環境, AMBIENCE\#GENERAL, 乾淨明亮, 7.00\#5.50)\\ \hline

\multirow{3}{*}{zho-lap} & ST 1 & 確實是一台功能強大的\underline{筆電}！\par (It is indeed a powerful laptop!) & 7.00\#6.38 \\ \cmidrule(lr){2-4}
& ST 2 & 這個保固說實在不是很完美。 \par (Honestly, this warranty is not very perfect.) & (保固, 不是很完美, 4.38\#5.25)\\ \cmidrule(lr){2-4}
& ST 3 & M16的外型真的是很好看，但是售價真的挺高，感謝分享\par (The M16 really looks great, but the price is quite high. Thanks for sharing.)& (M16的外型, LAPTOP\#DESIGN]\_FEATURES, 真的是很好看, 6.70\#6.70)\par (售價, LAPTOP\#PRICE, 真的挺高, 3.75\#6.12)\\ \hline

\multirow{1}{*}{zho-fin} & ST 1 & \underline{利息保障倍數}由53.16降至12.59。\par (The interest coverage ratio decreased from 53.16 to 12.59.) & 4.00\#4.83 \\

\bottomrule
\end{tabularx}
\end{table}

\section{Annotation Guidelines}
\label{sec:annotation-guidelines}

Aspect-Based Sentiment Analysis (ABSA) is a fine-grained sentiment analysis task that aims to identify what people are expressing sentiments about in a text (the aspect) and how they feel about it (the opinion). Unlike general sentiment classification, which assigns a single sentiment label to an entire text, ABSA decomposes sentiment into structured representations at the aspect level for more precise interpretation. \\
This annotation task involves extracting sentiment information as structured quadruplets, and further annotating Valence-Arousal Scores to replace the sentiment labels (Positive, Negative or Neutral). Annotators extract quadruplets of the form: \textbf{(Aspect Term, Aspect Category, Opinion Term, Valence\#Arousal)}. In this phase, you will identify four key components for each sentiment expressed:\\
\begin{enumerate}
    \item \textbf{Aspect Term (A)}: The specific word or phrase in the sentence that refers to an \textbf{attribute or component} of the \textbf{subject} being discussed.
    \item \textbf{Aspect Category (C)}: A predefined, \textbf{broad classification} that the aspect term falls under.
    \item \textbf{Opinion Term (O)}: The word or phrase that expresses sentiment (positive, negative, or neutral) towards the aspect term.
    \item \textbf{Valence (V)} and \textbf{Arousal (A)} are \textbf{continuous emotion dimensions} used to describe the emotional tone of an opinion expression beyond discrete labels like “positive” or “negative.”
\begin{itemize}
    \item \textbf{Valence}: Valence refers to the degree of positivity or negativity associated with an emotion or sentiment and can range from 1 (very negative) to 9 (very positive). Examples: "delightful" → \textbf{high valence} (e.g., 8.5); "horrible" → \textbf{low valence} (e.g., 2.0)
    \item \textbf{Arousal}: Arousal captures the intensity or activation level of the emotional state and can range from 1 (very calm/low energy) to 9 (very excited/high energy). Examples: "calm" → \textbf{low arousal} (e.g., 2.5); "furious" → \textbf{high arousal} (e.g., 8.5) 
\end{itemize}
\end{enumerate}

\subsection{Annotation Rules for Triplets}
\begin{table}[H]
\centering
\renewcommand{\arraystretch}{1.2} 
\begin{tabular}{l| p{13cm}} 
\toprule
\textbf{Rule 1} & There can be multiple aspect-opinion combinations in a sentence. Extract all satisfactory aspect-opinion combinations separately.\\ \cline{1-2}

 Example & Beautiful decor, great atmosphere, amazing food\\ \cline{1-2}
Triplets & \makecell[l]{(decor, RESTAURANT\#STYLE\_OPTIONS, Beautiful) \\ (atmosphere, AMBIENCE\#GENERAL’, great)\\ (food, FOOD\#QUALITY, amazing) } \\ \Xcline{1-2}{0.5pt}
\\[-3ex] 
\Xcline{1-2}{0.5pt}
\textbf{Rule 2} & If adverbs or negators are being used with an opinion, it should be included as well in the triplets because such modifiers may affect valence-arousal scores.\\ \cline{1-2}
 Example & I think the service was tremendously horrible but the food they served was absolutely delicious\\ \cline{1-2}
Triplets & \makecell[l]{(service, SERVICE\#GENERAL, tremendously horrible)  \\ (food, FOOD\#QUALITY, absolutely delicious)}\\ \Xcline{1-2}{0.5pt} 
\\[-3ex] 
\Xcline{1-2}{0.5pt} 
\textbf{Rule 3} & Don’t annotate words that aren’t even required. Keep it precise and only annotate what is required.\\ \cline{1-2}
 Example & I also ordered room service last night from this restaurant, and it was amazing.\\ \cline{1-2}
Triplets & \makecell[l]{(room service, SERVICE\#GENERAL, amazing)} \\ \Xcline{1-2}{0.5pt} 
\\[-3ex] 
\Xcline{1-2}{0.5pt} 
\textbf{Rule 4} & Use the words in the sentences. Do not add extra words to the annotations.\\ \cline{1-2}
Example & The staff is always friendly and helpful.\\ \cline{1-2}
Triplets & \makecell[l]{(staff, SERVICE\#GENERAL, always friendly) \\ (staff, SERVICE\#GENERAL, helpful) \\ Notes: The opinion is “helpful”, not “always helpful”} \\ \Xcline{1-2}{0.5pt} 
\\[-3ex] 
\Xcline{1-2}{0.5pt} 
\textbf{Rule 5} & Do not create new aspect categories. Only use the ones which are mentioned. Only use the following combinations, don’t use any other word as an aspect category. 
\\
\bottomrule
\end{tabular}
\end{table}

\subsection{Aspect Category List}
\begin{itemize}[leftmargin=*]
    \item Laptop
\vspace{-8pt}
\begin{table}[H]
\centering
\small
\setlength{\tabcolsep}{4pt}
\renewcommand{\arraystretch}{0.4}
    \begin{subtable}{1.0\textwidth}
        \centering
        \renewcommand{\arraystretch}{1.5} 
        \begin{tabularx}{\textwidth}{|>{\arraybackslash}X|}
            \hline
            \textbf{Entity Labels} \\ \hline
            LAPTOP, DISPLAY, KEYBOARD, MOUSE, MOTHERBOARD, CPU, FANS\_COOLING, PORTS, MEMORY, POWER\_SUPPLY, OPTICAL\_DRIVES, BATTERY, GRAPHICS, HARD\_DISK, MULTIMEDIA\_DEVICES, HARDWARE, SOFTWARE, OS, WARRANTY, SHIPPING, SUPPORT, COMPANY \\ \hline
            \textbf{Attribute Labels} \\ \hline
           GENERAL, PRICE, QUALITY, DESIGN\_FEATURES, OPERATION\_PERFORMANCE, USABILITY, PORTABILITY, CONNECTIVITY, MISCELLANEOUS \\ \hline
        \end{tabularx}
    \end{subtable}
\label{tab:Label Category}
\end{table}

\clearpage
    \item Restaurant
\vspace{-8pt}
\begin{table}[H]
\centering
\small
\setlength{\tabcolsep}{4pt}
\renewcommand{\arraystretch}{0.4}
\begin{subtable}[t]{1.0\textwidth}
        \centering
        \renewcommand{\arraystretch}{1.5} 
        \begin{tabularx}{\textwidth}{|>{\arraybackslash}X|}
            \hline
            \textbf{Entity Labels} \\ \hline
            RESTAURANT, FOOD, DRINKS, AMBIENCE, SERVICE, LOCATION \\ \hline
            \textbf{Attribute Labels} \\ \hline
            GENERAL, PRICES, QUALITY, STYLE\_OPTIONS, MISCELLANEOUS
 \\ \hline
        \end{tabularx}
\end{subtable}
\label{tab:Label Category}
\end{table}

\vspace{-20pt}
    \item Hotel
\vspace{-8pt}
\begin{table}[H]
\small
\centering
\setlength{\tabcolsep}{4pt}
\renewcommand{\arraystretch}{0.4}
    \begin{subtable}{1.0\textwidth}
        \centering
        \renewcommand{\arraystretch}{1.5} 
        \begin{tabularx}{\textwidth}{|>{\arraybackslash}X|}
            \hline
            \textbf{Entity Labels} \\ \hline
            HOTEL, ROOMS, FACILITIES, ROOM\_AMENITIES, SERVICE, LOCATION, FOOD\_DRINKS \\ \hline
            \textbf{Attribute Labels} \\ \hline
            GENERAL, PRICE, COMFORT, CLEANLINESS, QUALITY, DESIGN\_FEATURES, STYLE\_OPTIONS, MISCELLANEOUS \\ \hline
        \end{tabularx}
    \end{subtable}
\label{tab:Label Category}
\end{table}

\end{itemize}

\subsection{Combined Interpretation for Valence-Arousal Score}
The \textbf{Valence–Arousal} pair provides a more nuanced understanding of sentiment:
\begin{table}[H]
\centering
\renewcommand{\arraystretch}{1.25}

\begin{subtable}[t]{1.0\textwidth}
    \centering
    \begin{tabularx}{\textwidth}{|l|c|c|>{\raggedright\arraybackslash}X|}
        \hline
        \textbf{Expression} & \textbf{V} & \textbf{A} & \textbf{Interpretation} \\ \hline
        "very courteous"& 7.5 & 5.7 & (7.5, 5.7) Moderately positive and moderately active \\ \hline
        "terrible" & 2.8 & 6.8 & (2.8, 6.8) Strongly negative and emotionally intense \\ \hline
        "okay" & 5.0 & 3.0 & (5.0, 3.0) Neutral valence and low emotional intensity \\ \hline
    \end{tabularx}
\end{subtable}
\end{table}

The score distribution of valence and arousal often follows a \textbf{smile-shaped or U-shaped curve}.
\begin{itemize}
\item The higher the valence, the higher the arousal.
\item The lower the valence, the higher the arousal as well.
\item When valence is neutral, arousal tends to be lower.
\end{itemize}

\begin{figure}[ht]
    \centering
    \begin{subfigure}[tb]{0.85\textwidth}
        \centering
        \includegraphics[width=\textwidth]{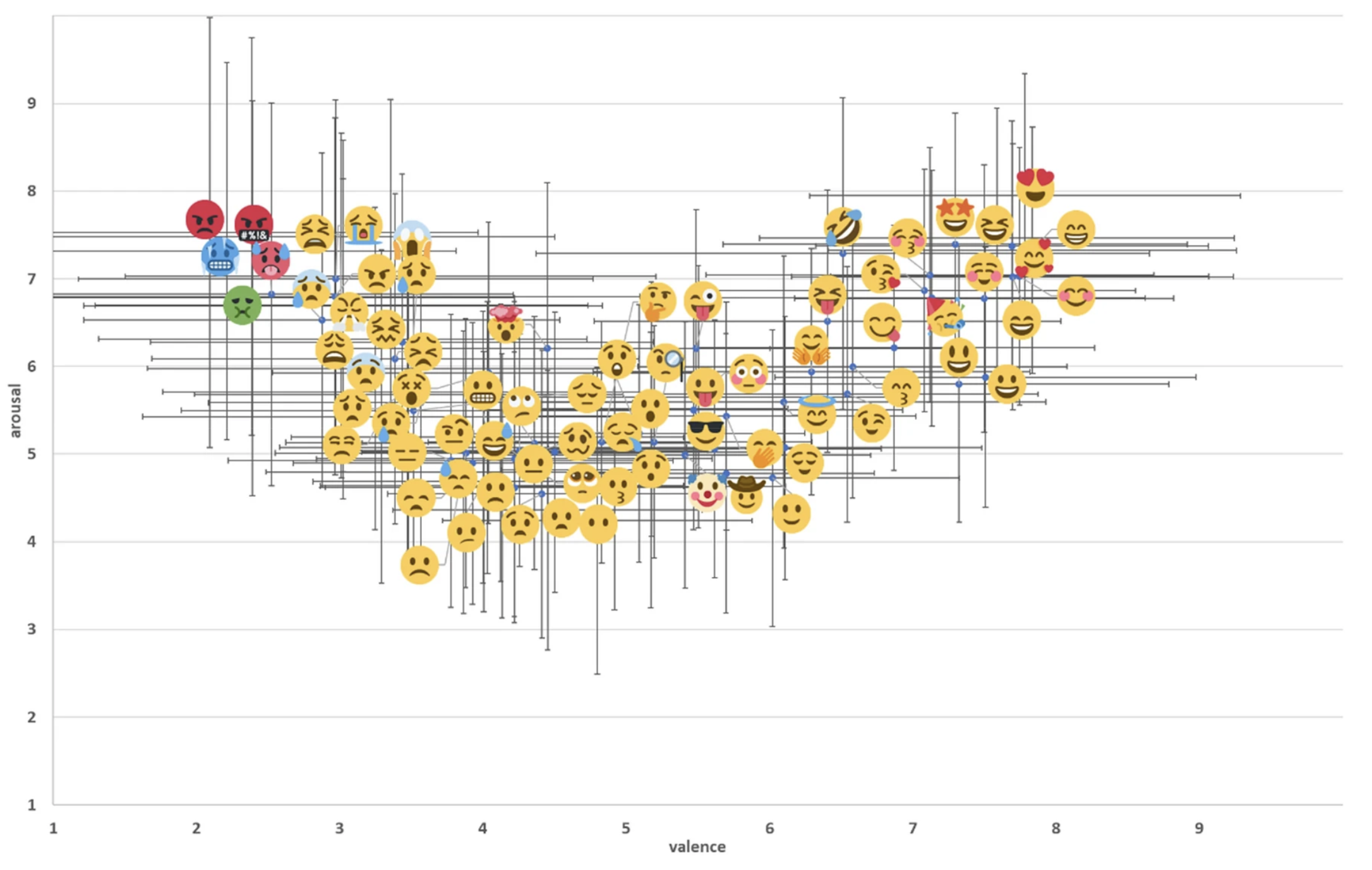}
    \end{subfigure}
\caption{Facial emoji\textquoteright s emotional states on the valence-arousal axes \cite{Kutsuzawa2022}.}
\end{figure}

\onecolumn
\subsection{Annotation Screens}
\vspace{-10pt}
\begin{figure*}[h]
    \centering

    \begin{subfigure}[b]{0.9\textwidth}
        \centering
        \includegraphics[width=\textwidth]{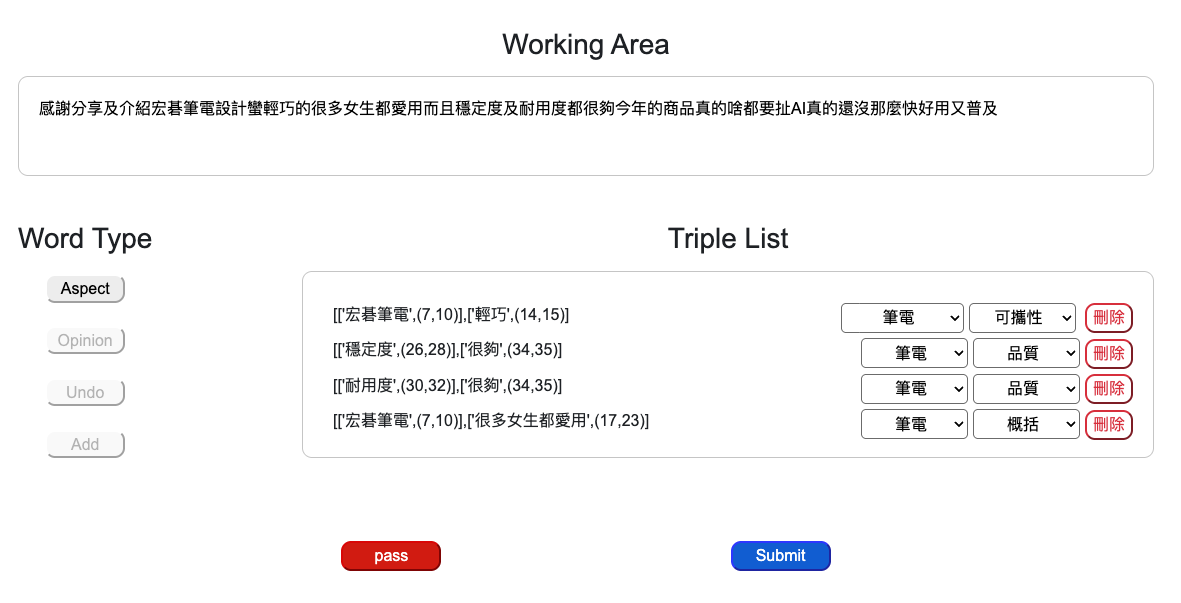}
        \caption{Triplet (A, C, O) annotation interface}
        \label{fig:aspect}
    \end{subfigure}
    
    \vspace{50pt}
    
    \begin{subfigure}[b]{0.9\textwidth}
        \centering
        \includegraphics[width=\textwidth]{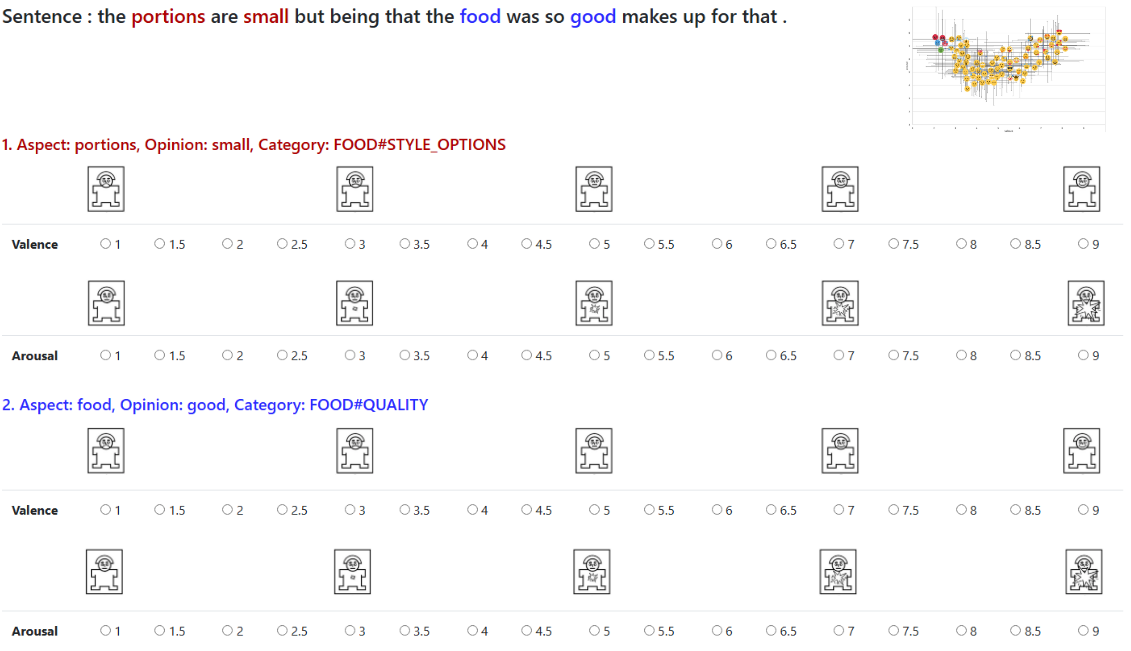}
        \caption{VA annotation interface}
        \label{fig:va}
    \end{subfigure}
    
\end{figure*}

\clearpage

\raggedbottom
\section{Category Distribution}
\label{longtail}
\begin{figure*}[h!]
    \centering

    \begin{subfigure}[b]{0.95\textwidth}
        \centering
        \includegraphics[width=\textwidth]{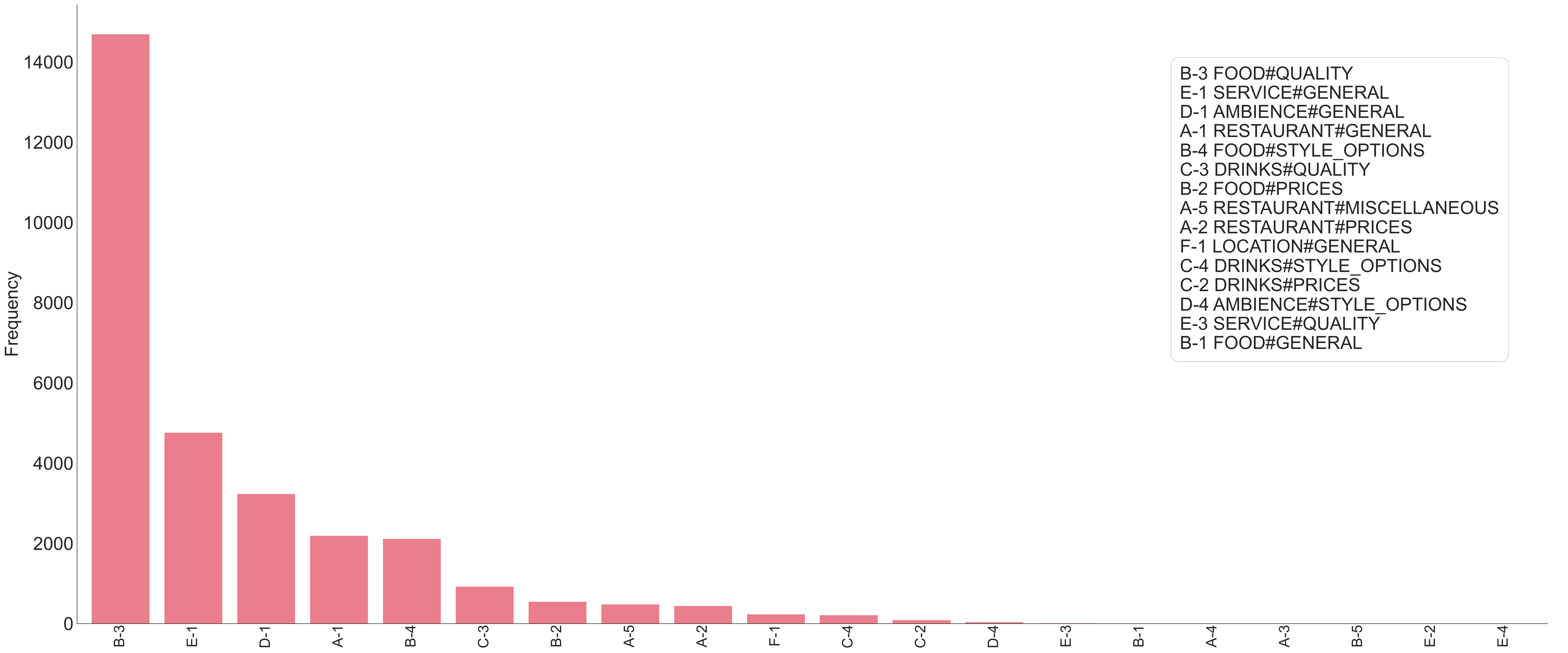}
        \caption{Restaurant Category Distribution}
        \label{fig:aspect}
    \end{subfigure}
    
    \vspace{30pt}

    \begin{subfigure}[b]{0.95\textwidth}
        \centering
        \includegraphics[width=\textwidth]{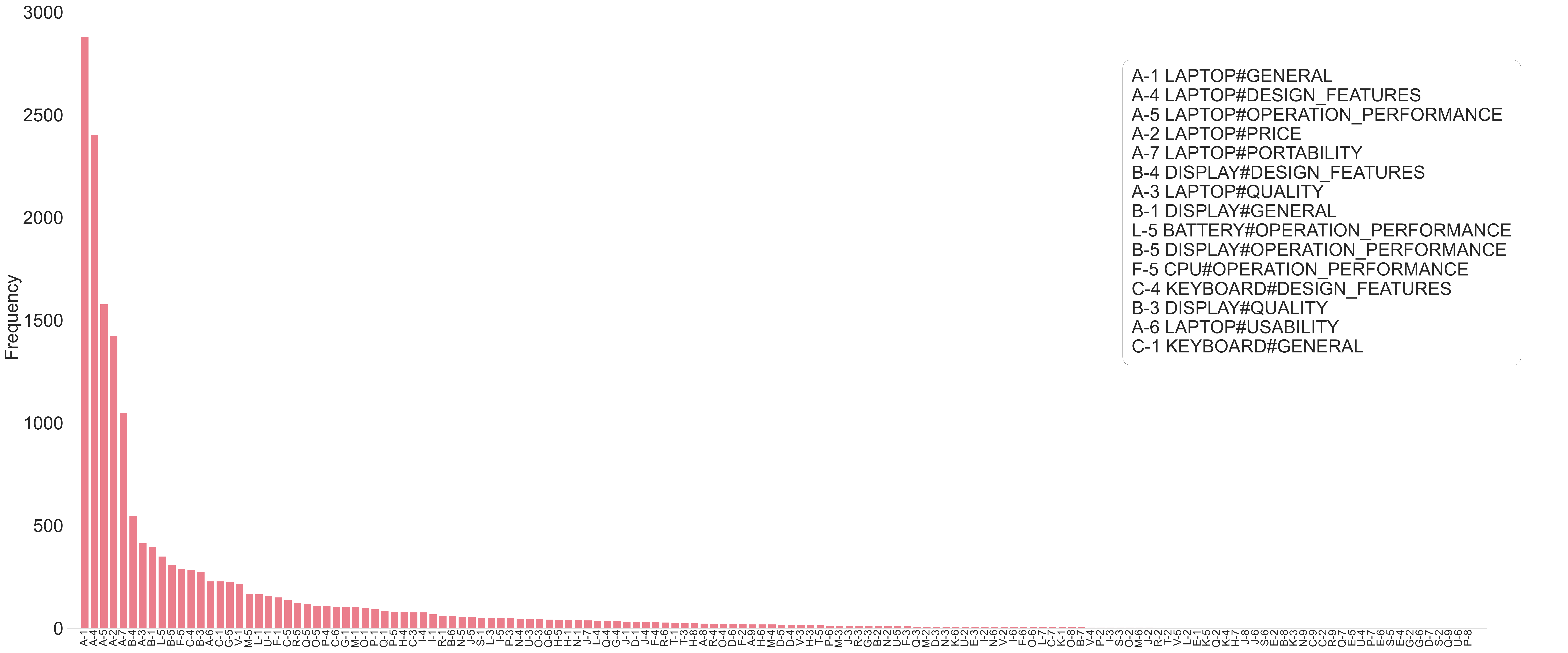}
        \caption{Laptop Category Distribution}
        \label{fig:aspect}
    \end{subfigure}
    
    \vspace{30pt}
    
    \begin{subfigure}[b]{0.95\textwidth}
        \centering
        \includegraphics[width=\textwidth]{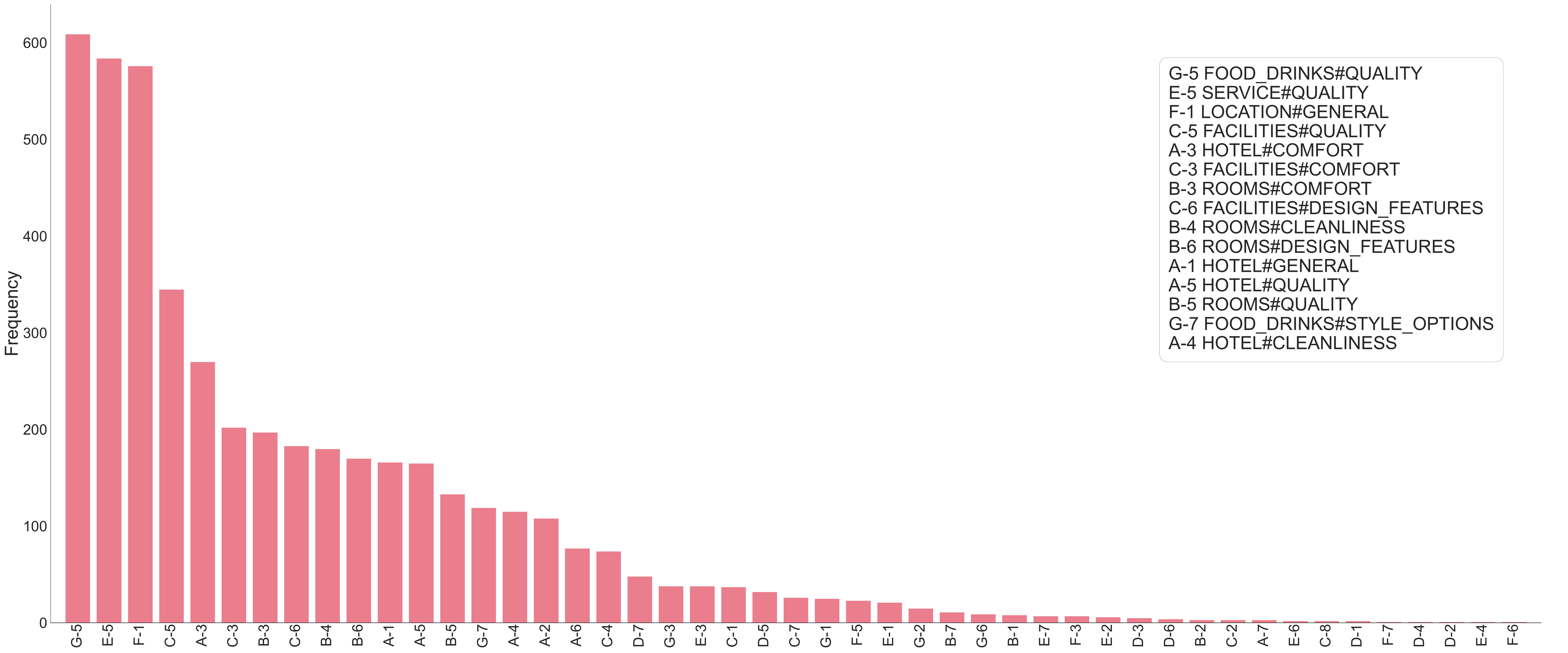}
        \caption{Hotel Category Distribution}
        \label{fig:va}
    \end{subfigure}
    
\end{figure*}

\section{Prompts used for Few-Shot Learning}
\label{sec:fewshot_prompt}
\begin{table}[H]
\centering
\setlength{\tabcolsep}{4pt}
    \begin{subtable}{1.0\textwidth}
     
        \centering
        \renewcommand{\arraystretch}{1.2}
        \keepXColumns
        \begin{tabularx}{\textwidth}{|>{\arraybackslash}X|}
            \hline
            \textbf{DimASR} \\ \hline
            Definition: The output will be a pair of real numbers [valence\#arousal] representing the sentiment toward the identified aspect in the sentence, where both values are on a scale from 1 to 9.\\[-0.3em]
            Valence: 1 = most negative, 9 = most positive;\\[-0.3em]
            Arousal: 1 = calm/low intensity, 9 = excited/high intensity.\\[-0.3em]
            \textbf{<Examples>}\\[-0.3em]
            Now complete the following example\\[-0.3em]
            \textbf{Input:} <sentence> The aspect is <aspect>.\\[-0.3em]
            \textbf{Output:}\\ \hline
            \textbf{DimASQP} \\[-0.3em](Notes: without “aspect category” and its corresponding part is used for dimASTE)\\ \hline
            According to the following sentiment elements definition:\\[-0.3em]
            - The “aspect term” refers to a specific feature, attribute, or aspect of a product or service on which a user can express an opinion. Explicit aspect terms appear explicitly as a substring of the given text.\\[-0.3em]
            - The "aspect category" refers to the category that aspect belongs to, and the available categories include: \textbf{<Aspect\_Category\_Lists>}\\[-0.3em]
            - The “opinion term” refers to the sentiment or attitude expressed by a user towards a particular aspect or feature of a product or service. Explicit opinion terms appear explicitly as a substring of the given text.\\[-0.3em]
            - The “sentiment score” refers to the emotional response toward the aspect, represented as a Valence-Arousal pair, where Valence indicates positivity/negativity (1.0 = most negative, 9.0 = most positive) and Arousal indicates emotional intensity (1.0 = calm, 9.0 = excited). Both values are real numbers in the range [1.0, 9.0].\\[-0.3em]
            Please carefully follow the instructions. Ensure that aspect terms are recognized as exact matches in the review. Ensure that opinion terms are recognized as exact matches in the review.\\[-0.3em]
            Recognize all sentiment elements with their corresponding aspect terms, aspect categories, opinion terms, and sentiment score in the given input text (review). Provide your response in the format of a Python list of tuples: 'Sentiment elements: [("aspect term", "aspect category", "opinion term", [valence\#arousal]), ...]'. Note that “, ...” indicates that there might be more tuples in the list if applicable and must not occur in the answer. Ensure there is no additional text in the response.\\[-0.3em]
            \textbf{<Examples>}\\[-0.3em]
            \textbf{Input:} <sentence>\\[-0.3em]
            \textbf{Output:}\\ \hline
        \end{tabularx}
    \end{subtable}
\label{tab:Label Category}
\end{table}

\section{Prompts used for Supervised Fine-Tuning}
\label{sec:sft_prompt}
\begin{table}[H]
\centering
\setlength{\tabcolsep}{4pt}
\renewcommand{\arraystretch}{1.25}
    \begin{subtable}{1.0\textwidth}
     
        \centering
        \renewcommand{\arraystretch}{1.5}
        \keepXColumns
        \begin{tabularx}{\textwidth}{|>{\arraybackslash}X|}
            \hline
            \textbf{DimASR} \\ \hline
            \textbf{System Prompt}:\\
            Instruction: The output will be a pair of real numbers [valence\#arousal] representing the sentiment toward the identified aspect in the sentence, where both values are on a scale from 1 to 9. Valence: 1 = most negative, 9 = most positive; Arousal: 1 = calm/low intensity, 9 = excited/high intensity.\\
            Provide your response in the format of a Python list: [valence\#arousal]\\
            \textbf{User Prompt}:\\
            Text: <text> \\
            Aspect Term: <aspect>\\
            \textbf{Assistant Prompt:}\\
            < valence\#arousal >\\ \hline
            \textbf{DimASQP} \\(Notes: without “aspect category” and its corresponding part is used for DimASTE)\\ \hline
            \textbf{System Prompt:}\\
            Instruction: According to the following sentiment elements definition.
            - The “aspect term” refers to a specific feature, attribute, or aspect of a product or service on which a user can express an opinion. Explicit aspect terms appear explicitly as a substring of the given text.\\
            - The "aspect category" refers to the category that aspect belongs to, and the available categories include: \textbf{<Aspect\_Category\_Lists>}\\
            - The “opinion term” refers to the sentiment or attitude expressed by a user towards a particular aspect or feature of a product or service. Explicit opinion terms appear explicitly as a substring of the given text.\\
            - The “sentiment score” refers to the emotional response toward the aspect, represented as a Valence-Arousal pair, where Valence indicates positivity/negativity (1.0 = most negative, 9.0 = most positive) and Arousal indicates emotional intensity (1.0 = calm, 9.0 = excited). Both values are real numbers in the range [1.0, 9.0].\\
            Please carefully follow the instructions. Ensure that aspect terms are recognized as exact matches in the review. Ensure that opinion terms are recognized as exact matches in the review.\\
            \textbf{User Prompt:}\\
            Text: <text> \\
            \textbf{Assistant Prompt:}\\
            Sentiment elements: <tuple>\\ \hline
        \end{tabularx}
    \end{subtable}
\label{tab:Label Category}
\end{table}

\clearpage
\raggedbottom
\section{Example Calculation of cF1}
\label{sec:example_calculating_cF1}

\begin{table}[H]
\centering
\renewcommand{\arraystretch}{1.1}
\begin{tabular*}{\textwidth}{@{\extracolsep{\fill}}|l|c|c|c|c|}
\hline

\multirow{2}{*} &  & \multicolumn{2}{c|}{\textbf{VA error distance}} & \\ \cline{3-4}
 {\textbf{Prediction/Gold}} &\multicolumn{1}{p{1.5cm}|}{\centering $TP_{cat}$ \\ (A)} & \multicolumn{1}{p{1.5cm}|}{\centering Raw \\ (B)} & \multicolumn{1}{p{3cm}|}{\centering Normalized \\ $(C)=(B)/ \sqrt{128}$} & \multicolumn{1}{p{1.5cm}|}{\centering \textbf{cTP} \\ (A)-(C)}\\ \hline
 
P: (food, good, 8.00\#8.00) & \multirow{2}{*}{1} & \multirow{2}{*}{$\sqrt{2}$} & \multirow{2}{*}{$\frac{\sqrt{2}}{\sqrt{128}}= 0.125$} & \multirow{2}{*}{0.875} \\
G: (food, good, 7.00\#7.00) & & & & \\ \hline

P: (soup, spicy, 7.50\#7.50) & \multirow{2}{*}{1} & \multirow{2}{*}{$\sqrt{32}$} & \multirow{2}{*}{$\dfrac{\sqrt{32}}{\sqrt{128}} = 0.5$} & \multirow{2}{*}{0.5} \\
G: (soup, spicy, 3.50\#3.50) & & & & \\ \hline

P: (staff, friendly, 7.00\#7.00) & \multirow{2}{*}{0} & \multirow{2}{*}{-} & \multirow{2}{*}{-} & \multirow{2}{*}{0} \\
G: (staff, always friendly, 7.50\#7.50) & & & & \\ \hline

P: (staff, good, 7.00\#7.00) & \multirow{2}{*}{0} & \multirow{2}{*}{-} & \multirow{2}{*}{-} & \multirow{2}{*}{0} \\
G: N/A & & & & \\ \hline
\multicolumn{3}{|c|}{} & Total cTP& 1.375 \\ \hline
\multicolumn{5}{|r|}{$cRecall = 1.375/3 = 0.458$} \\ \hline
\multicolumn{5}{|r|}{$cPrecision = 1.375/4 = 0.344$}  \\ \hline
\multicolumn{5}{|r|}{$cF1= (2*0.458*0.344)/(0.458+0.344)=0.393$}  \\ \hline
\end{tabular*}
\end{table}

Note: The VA scores lie in the range [1, 9]. When the VA prediction is perfect (i.e., dist=0), cRecall/cPrecision reduces to the standard recall/precision. 

\clearpage

\section{DimASTE and DimASQP Results}
\label{DimASTE-ASQPResults}
\begin{table}[H]
\centering
\small
\setlength{\tabcolsep}{2.2pt}
\renewcommand{\arraystretch}{1.0}

\newcommand{\best}[1]{\textbf{#1}}
\begin{tabular*}{\textwidth}{@{\extracolsep{\fill}}ll ccc|ccc|ccc|ccc}
\toprule
&& \multicolumn{6}{c|}{Zero-Shot Learning} 
& \multicolumn{6}{c}{One-Shot Learning} \\
\cmidrule(lr){3-8} \cmidrule(lr){9-14}
Subtask&Dataset & \multicolumn{3}{c|}{\centering GPT-5 mini} & \multicolumn{3}{c|}{Kimi K2 Thinking} & \multicolumn{3}{c|}{\centering GPT-5 mini} & \multicolumn{3}{c}{ Kimi K2 Thinking}\\ 
\cmidrule(lr){3-5}\cmidrule(lr){6-8} \cmidrule(lr){9-11}\cmidrule(lr){12-14}
&& cP & cR & cF1 & cP & cR & cF1 & cP & cR & cF1 & cP & cR & cF1\\ \midrule

\multirow{8}{*}{DimASTE} & eng-rest & 0.4719&0.5301&0.4993 & 0.5043 & 0.5159&\best{0.5101}&0.4667&0.5465&\best{0.5034}&0.4902&0.4938&0.4920\\
&eng-lap & 0.4224 & 0.4793 & 0.4491 & 0.4458 & 0.4582 & \best{0.4519} & 0.4443 & 0.5399 & \best{0.4874} & 0.4265 & 0.4595 & 0.4424\\
&jpn-hot & 0.1552 & 0.1947 & 0.1727 & 0.2975 & 0.3342 & \best{0.3148} &  0.2259 & 0.2767 & 0.2487 & 0.3409 & 0.3520 & \best{0.3464}\\
&rus-rest & 0.3575 & 0.4580 & 0.4016 & 0.3863 & 0.4214 & \best{0.4031} & 0.3380 & 0.4159 & 0.3730 & 0.4287 & 0.4199 & \best{0.4242}\\
&tat-rest & 0.3038 & 0.3910 & 0.3419 & 0.3363 & 0.3656 & \best{0.3503} & 0.2868 & 0.3516 & 0.3159 & 0.3594 & 0.3561 & \best{0.3577} \\
&ukr-rest & 0.3571 & 0.4583 & 0.4014 & 0.3875 & 0.4313 & \best{0.4082} & 0.3164 & 0.3505 & 0.3326 & 0.4229 & 0.4212 & \best{0.4220} \\
&zho-rest & 0.2737& 0.3823& 0.3190 & 0.3436 & 0.4076 & \best{0.3728} & 0.2115 & 0.2842 & 0.2425 & 0.3357 & 0.3719 & \best{0.3529} \\
&zho-lap & 0.1934 & 0.3065 & \best{0.2372} & 0.1975 & 0.2553 & 0.2227 & 0.2271 & 0.3702 & \best{0.2815} & 0.2233 & 0.2823 & 0.2494 \\
\cmidrule(lr){1-14}
&AVG & 0.3169 & 0.4000 & 0.3528 & 0.3624 & 0.3987& \best{0.3792}& 0.3146 & 0.3919 & 0.3481& 0.3785 & 0.3946 & \best{0.3859}\\
\cmidrule(lr){1-14}\morecmidrules\cmidrule(lr){1-14}

\multirow{8}{*}{DimASQP} & eng-rest & 0.3814 & 0.4285 & \best{0.4036} & 0.3698 & 0.3783 & 0.3740 & 0.3538 & 0.4143 & \best{0.3816} & 0.3732 & 0.3760 & 0.3746 \\
&eng-lap & 0.2167 & 0.2459 & 0.2304 & 0.2767 & 0.2844 & \best{0.2805} & 0.2590 & 0.3148 & \best{0.2842} & 0.2695 & 0.2903 & 0.2795 \\
&jpn-hot & 0.0815 & 0.1022 & 0.0907 & 0.1237 & 0.1389 & \best{0.1309} & 0.1400 & 0.1715 & 0.1542 & 0.1912 & 0.1974 & \best{0.1943} \\
&rus-rest & 0.2233 & 0.2860 & 0.2508 & 0.2546 & 0.2777 & \best{0.2656} & 0.2271 & 0.2794 & 0.2505 & 0.2994 & 0.2933 & \best{0.2963}\\
&tat-rest & 0.1754 & 0.2257 & 0.1974 & 0.2217 & 0.2410 & \best{0.2310} & 0.1839 & 0.2254 & 0.2025 & 0.2391 & 0.2369 & \best{0.2380} \\
&ukr-rest & 0.2193 & 0.2814 & 0.2465 & 0.2823 & 0.3142 & \best{0.2974} & 0.2074 & 0.2297 & 0.2180 & 0.2977 & 0.2966 & \best{0.2971} \\
&zho-rest & 0.2129 & 0.2974 & 0.2481 & 0.2741 & 0.3252 & \best{0.2975} & 0.1649 & 0.2217 & 0.1891 & 0.2720 & 0.3013 & \best{0.2859} \\
&zho-lap & 0.1105 & 0.1752 & 0.1356 & 0.1391 & 0.1799 & \best{0.1569} & 0.1549 & 0.2526 & \best{0.1921} & 0.1701 & 0.2150 & 0.1900 \\
\cmidrule(lr){1-14}
&AVG & 0.2026 & 0.2553 & 0.2254 & 0.2428 & 0.2675 & \best{0.2542} & 0.2114 & 0.2637 & 0.2340 & 0.2640 & 0.2759 & \best{0.2695} \\
\bottomrule
\end{tabular*}

\vspace{0.6em}

\begin{tabular*}{\textwidth}{@{\extracolsep{\fill}}ll ccc|ccc|ccc|ccc}
\toprule
&& \multicolumn{12}{c}{Supervised Fine-Tuning }\\
\cmidrule(lr){3-14}
Subtask&Dataset & \multicolumn{3}{c|}{\centering Qwen3 (14B)} & \multicolumn{3}{c|}{Ministral-3 (14B)} & \multicolumn{3}{c|}{\centering Llama-3.3 (70B)} & \multicolumn{3}{c}{ GPT-OSS (120B)}\\ 
\cmidrule(lr){3-5}\cmidrule(lr){6-8} \cmidrule(lr){9-11}\cmidrule(lr){12-14}
&& cP & cR & cF1 & cP & cR & cF1 & cP & cR & cF1 & cP & cR & cF1\\ \midrule

\multirow{8}{*}{DimASTE} & eng-rest & 0.4634 & 0.4341 & 0.4483 & 0.2913 & 0.2948 & 0.2930 & 0.5547 & 0.5294 & 0.5418 & 0.5644 & 0.5254 & \best{0.5442}\\
&eng-lap & 0.3944 & 0.3716 & 0.3827 & 0.2673 & 0.2803 & 0.2736 & 0.4939 & 0.4418 & \best{0.4664} & 0.4854 & 0.4220 & 0.4515\\
&jpn-hot & 0.1685 & 0.1563 & 0.1622 & 0.1404 & 0.1517 & 0.1458 & 0.4663 & 0.4724 & 0.4694 & 0.5549 & 0.5253 & \best{0.5397}\\
&rus-rest & 0.3310 & 0.3373 & 0.3341 & 0.1742 & 0.1808 & 0.1774 & 0.4454 & 0.4736 & \best{0.4590} & 0.4217 & 0.4307 & 0.4262\\
&tat-rest & 0.2033 & 0.2008 & 0.2020 & 0.1170 & 0.1139 & 0.1154 & 0.3946 & 0.4268 & \best{0.4101} & 0.3406 & 0.3768 & 0.3578 \\
&ukr-rest & 0.3077 & 0.3121 & 0.3099 & 0.1555 & 0.1637 & 0.1595 & 0.4466 & 0.4568 & \best{0.4517} & 0.4189 & 0.4314 & 0.4250 \\
&zho-rest & 0.2528 & 0.2490 & 0.2509 & 0.1373 & 0.1527 & 0.1446 & 0.4837 & 0.4741 & \best{0.4789} & 0.4674 & 0.4848 & 0.4759 \\
&zho-lap & 0.2021 & 0.2182 & 0.2099 & 0.1052 & 0.1348 & 0.1182 & 0.4432 & 0.4260 & 0.4344 & 0.4468 & 0.4268 & \best{0.4366} \\
\cmidrule(lr){1-14}
&AVG & 0.2904 & 0.2849 & 0.2875 & 0.1735 & 0.1841 & 0.1784 & 0.4661 & 0.4626 & \best{0.4640} & 0.4625 & 0.4529 & 0.4571\\
\cmidrule(lr){1-14}\morecmidrules\cmidrule(lr){1-14}

\multirow{8}{*}{DimASQP} & eng-rest & 0.2763 & 0.2588 & 0.2673 & 0.2046 & 0.2070 & 0.2058 & 0.5169 & 0.4933 & \best{0.5048} & 0.5199 & 0.4840 & 0.5013 \\
&eng-lap & 0.1576 & 0.1485 & 0.1529 & 0.1263 & 0.1325 & 0.1293 & 0.2630 & 0.2353 & \best{0.2483} & 0.2592 & 0.2253 & 0.2411 \\
&jpn-hot & 0.0416 & 0.0386 & 0.0400 & 0.0299 & 0.0323 & 0.0311 & 0.3554 & 0.3601 & 0.3577 & 0.4268 & 0.4040 & \best{0.4151} \\
&rus-rest & 0.1666 & 0.1698 & 0.1682 & 0.0981 & 0.1019 & 0.1000 & 0.3995 & 0.4248 & \best{0.4118} & 0.3644 & 0.3722 & 0.3683\\
&tat-rest & 0.0959 & 0.0948 & 0.0954 & 0.0619 & 0.0603 & 0.0611 & 0.3562 & 0.3853 & \best{0.3702} & 0.2945 & 0.3258 & 0.3094 \\
&ukr-rest & 0.1631 & 0.1653 & 0.1641 & 0.0951 & 0.1001 & 0.0975 & 0.4024 & 0.4117 & \best{0.4070} & 0.3610 & 0.3718 & 0.3663 \\
&zho-rest & 0.1617 & 0.1593 & 0.1605 & 0.0887 & 0.0986 & 0.0934 & 0.4436 & 0.4348 & \best{0.4391} & 0.4173 & 0.4329 & 0.4249\\
&zho-lap & 0.1083 & 0.1169 & 0.1124 & 0.0648 & 0.0831 & 0.0728 & 0.3577 & 0.3438 & 0.3506 & 0.3634 & 0.3471 & \best{0.3551} \\
\cmidrule(lr){1-14}
&AVG & 0.1464 & 0.1440 & 0.1451 & 0.0962 & 0.1020 & 0.0989 & 0.3868 & 0.3961 & \best{0.3862} & 0.3758 & 0.3704 & 0.3727 \\
\bottomrule
\end{tabular*}

\label{}
\end{table}

\clearpage

\section{Few-Shot Results}
\label{Few-shotResults}

\begin{table}[H]
\centering
\small
\setlength{\tabcolsep}{5pt}
\renewcommand{\arraystretch}{1.0}

\newcommand{\best}[1]{\textbf{#1}}
\begin{tabular*}{\textwidth}{@{\extracolsep{\fill}}ll cccccccccc}
\toprule
&& \multicolumn{10}{c}{Shots} \\
\cmidrule(lr){3-12}
Subtask&Dataset & 0 & 1 & 2 & 4 & 8 & 16 & 32 & 64 & 128 & 256 \\\midrule
\multirow{10}{*}{DimASR} & eng-rest & 2.9490&2.3926&2.1832&2.1351&1.7992&1.5631&1.4263&1.3036&1.2443&\best{1.2224}\\
&eng-lap &3.2115&2.5637&2.5269&2.0482&2.1466&1.9426&1.5950&1.3792&\best{1.3251}&1.3383 \\
&jpn-hot & 3.1406&2.1607&2.0899&2.0094&1.1898&0.8968&0.8478&0.8047&0.7924&\best{0.7498}\\
&jpn-fin & 2.6760&1.9243&1.5710&1.6804&1.2386&1.1088&1.1425&1.0645&1.0192&\best{0.9380}\\
&rus-rest & 2.5447&2.0390&2.0192&2.0296&1.8490&1.7170&1.5912&\best{1.5740}&1.6682&1.6385\\
&tat-rest & 2.6645&2.2308&2.2187&2.2111&1.9360&1.8583&\best{1.6913}&1.6990&1.7692&1.7318\\
&ukr-rest & 2.5628&2.0438&2.0222&2.0141&1.8340&1.7152&1.6259&\best{1.5723}&1.6515&1.6383\\
&zho-rest & 2.7125&2.2467&1.8937&1.9228&1.5810&1.3957&1.2485&1.2191&1.1436&\best{1.1169}\\
&zho-lap & 2.4790&1.9380&1.7699&1.4945&1.3179&1.0741&0.9765&0.9144&0.8802&\best{0.8781}\\
&zho-fin & 2.6547&2.0094&1.6099&1.3295&1.1311&0.8401&0.7343&0.7253&0.6825&\best{0.6532}\\
\cmidrule(lr){1-12}
&AVG & 2.7595&2.1549&1.9904&1.8875&1.6023&1.4112&1.2879&1.2256&1.2176&\best{1.1905}\\
\cmidrule(lr){1-12}\morecmidrules\cmidrule(lr){1-12}
\multirow{8}{*}{DimASTE} & eng-rest & 0.4993&0.5034&0.5117&0.5234&0.5425&0.607&0.6201&\best{0.6274}&0.6212&0.6247 \\
&eng-lap & 0.4491&0.4874&0.4936&0.5210&0.5259&0.5302&0.5664&0.5646&0.5758&\best{0.5769} \\
&jpn-hot & 0.1727&0.2487&0.2721&0.2923&0.2975&0.3078&0.3204&0.3120&0.3085&\best{0.3261}\\
&rus-rest & 0.4016&0.3730&0.3466&0.4037&0.4375&0.4295&0.4247&\best{0.4360}&0.4217&0.4220 \\
&tat-rest & 0.3419&0.3159&0.3120&0.3599&0.3901&0.4055&0.3789&0.3779&\best{0.3823}&0.3677 \\
&ukr-rest & 0.4014&0.3326&0.3152&0.3834&\best{0.4117}&0.3994&0.3973&0.4081&0.3951&0.3890 \\
&zho-rest & 0.3190&0.2425&0.3166&0.3403&0.3705&0.3862&\best{0.4091}&0.4033&0.4066&0.3995 \\
&zho-lap & 0.2372&0.2815&0.2787&0.2785&0.2835&0.2758&0.2851&0.2891&0.2905&\best{0.2929} \\
\cmidrule(lr){1-12}
&AVG & 0.3528&0.3481&0.3558&0.3878&0.4074&0.4177&0.4253&\best{0.4273}&0.4252&0.4249 \\
\cmidrule(lr){1-12}\morecmidrules\cmidrule(lr){1-12}
\multirow{8}{*}{DimASQP} & eng-rest &0.4036&0.3816&0.4290&0.4585&0.4767&0.5453&0.5525&0.5689&0.5634&\best{0.5749} \\
&eng-lap & 0.2304&0.2842&0.3038&0.3170&0.3193&0.3255&0.3490&\best{0.3667}&0.3640&0.3473 \\
&jpn-hot & 0.0907&0.1542&0.1720&0.1912&0.1946&0.2032&\best{0.2301}&0.2251&0.2176&0.2264 \\
&rus-rest & 0.2508&0.2505&0.2496&0.3187&0.3516&0.3679&0.3726&\best{0.3977}&0.3861&0.3878\\
&tat-rest & 0.1974&0.2025&0.2049&0.2748&0.3081&\best{0.3494}&0.3311&0.3413&0.3437&0.3328 \\
&ukr-rest & 0.2465&0.2180&0.2168&0.2878&0.3234&0.3366&0.3441&\best{0.3644}&0.3561&0.3489\\
&zho-rest & 0.2481&0.1891&0.2529&0.2760&0.3172&0.3381&0.3685&0.3688&\best{0.3744}&0.3694 \\
&zho-lap & 0.1356&0.1921&0.2001&0.2051&0.2057&0.2068&0.2118&0.2192&0.2219&\best{0.2313}\\
\cmidrule(lr){1-12}
&AVG & 0.2254&0.2340&0.2536&0.2911&0.3121&0.3341&0.3450&\best{0.3565}&0.3534&0.3524 \\
\bottomrule
\end{tabular*}
\end{table}

\clearpage

\onecolumn
\section{Few-shot VA scatter}
\label{Few-shotVAscatter}

\begin{figure*}[h!]
    \centering
    \par\vspace{1em}
    \begin{subfigure}[b]{0.87\textwidth}
        \centering
        \includegraphics[width=\textwidth]{figures/va_scatter_few-shot/dimasr_eng_restaurant_va_scatter.png}
        \caption{eng-rest}
    \end{subfigure}
    \par\vspace{2.5em}
    \begin{subfigure}[b]{0.87\textwidth}
        \centering
        \includegraphics[width=\textwidth]{figures/va_scatter_few-shot/dimasr_eng_laptop_va_scatter.png}
        \caption{eng-lap}
    \end{subfigure}
    \par\vspace{2.5em}
    \begin{subfigure}[b]{0.87\textwidth}
        \centering
        \includegraphics[width=\textwidth]{figures/va_scatter_few-shot/dimasr_jpn_hotel_va_scatter.png}
        \caption{jpn-hot}
    \end{subfigure}
    \par\vspace{2.5em}
    \begin{subfigure}[b]{0.87\textwidth}
        \centering
        \includegraphics[width=\textwidth]{figures/va_scatter_few-shot/dimasr_jpn_finance_va_scatter.png}
        \caption{jpn-fin}
    \end{subfigure}
    \par\vspace{2.5em}
    \begin{subfigure}[b]{0.87\textwidth}
        \centering
        \includegraphics[width=\textwidth]{figures/va_scatter_few-shot/dimasr_rus_restaurant_va_scatter.png}
        \caption{rus-rest}
    \end{subfigure}

    \label{fig:DimASR_few-shot_va_scatter_1}
\end{figure*}

\begin{figure*}[h]\ContinuedFloat
    \centering
    \begin{subfigure}[b]{0.87\textwidth}
        \centering
        \includegraphics[width=\textwidth]{figures/va_scatter_few-shot/dimasr_tat_restaurant_va_scatter.png}
        \caption{tat-rest}
    \end{subfigure}
    \par\vspace{2.5em}
    \begin{subfigure}[b]{0.87\textwidth}
        \centering
        \includegraphics[width=\textwidth]{figures/va_scatter_few-shot/dimasr_ukr_restaurant_va_scatter.png}
        \caption{ukr-rest}
    \end{subfigure}
    \par\vspace{2.5em}
    \begin{subfigure}[b]{0.87\textwidth}
        \centering
        \includegraphics[width=\textwidth]{figures/va_scatter_few-shot/dimasr_zho_restaurant_va_scatter.png}
        \caption{zho-rest}
    \end{subfigure}
    \par\vspace{2.5em}
    \begin{subfigure}[b]{0.87\textwidth}
        \centering
        \includegraphics[width=\textwidth]{figures/va_scatter_few-shot/dimasr_zho_laptop_va_scatter.png}
        \caption{zho-lap}
    \end{subfigure}
    \par\vspace{2.5em}
    \begin{subfigure}[b]{0.87\textwidth}
        \centering
        \includegraphics[width=\textwidth]{figures/va_scatter_few-shot/dimasr_zho_finance_va_scatter.png}
        \caption{zho-fin}
    \end{subfigure}
    \par\vspace{2.5em}
    \label{fig:DimASR_few-shot_va_scatter_2}
\end{figure*}

\clearpage
\section{Categorical ABSA}
\label{Catgorical Results}
\subsection{Data Statistics}
 We split up all data in positive (V>5.5), neutral (4.5<=V<=5.5), and negative (V<4.5) samples. We report the average valence and arousal values for each split of the data. Counts (n) and standard deviations (SD) are shown in tiny.
\newcommand{\msd}{\textsubscript{\text{mean}}\,{\tiny(SD)}}

\begin{table}[H]
\centering
\small
\setlength{\tabcolsep}{2pt}
\renewcommand{\arraystretch}{1.25}

\begin{tabular}{l ccc ccc ccc}
\toprule
\multicolumn{1}{c}{DimABSA} 
& \multicolumn{3}{c}{Positive ($V>5.5$)} 
& \multicolumn{3}{c}{Neutral ($4.5 \le V \le 5.5$)} 
& \multicolumn{3}{c}{Negative ($V<4.5$)} \\
\cmidrule(lr){1-1} \cmidrule(lr){2-4} \cmidrule(lr){5-7} \cmidrule(lr){8-10}

Dataset& \%(n) & V\msd & A\msd
& \%(n) & V\msd & A\msd
& \%(n) & V\msd & A\msd \\
\midrule
eng-rest & 71.14\% {\tiny(5,720)} & 7.327 {\tiny(0.645)} & 7.241 {\tiny(0.692)}
& 9.15\%  {\tiny(736)}  & 5.016 {\tiny(0.290)} & 5.220 {\tiny(0.508)}
& 19.7\% {\tiny(1,584)}  & 3.007 {\tiny(0.804)} & 6.444 {\tiny(1.083)} \\
\cmidrule(lr){1-10}
eng-lap & 65.58\% {\tiny(6,401)} & 7.215 {\tiny(0.625)} & 7.151 {\tiny(0.670)}
& 9.42\%  {\tiny(919)}  & 4.948 {\tiny(0.320)} & 5.269 {\tiny(0.485)}
& 25.00\% {\tiny(2,440)}  & 3.234 {\tiny(0.720)} & 6.284 {\tiny(0.967)} \\
\midrule
jpn-hot & 83.46\% {\tiny(5,032)} & 6.803 {\tiny(0.464)} & 6.486 {\tiny(0.559)}
& 1.19\%  {\tiny(72)}  & 5.089 {\tiny(0.332)} & 5.418 {\tiny(0.581)}
& 15.34\% {\tiny(925)}  & 3.498 {\tiny(0.494)} & 6.015 {\tiny(0.555)} \\
\cmidrule(lr){1-10}
jpn-fin
& 59.10\% {\tiny(1,946)} & 6.189 {\tiny(0.426)} & 5.497 {\tiny(0.497)}
& 5.25\%  {\tiny(173)}  & 5.130 {\tiny(0.342)} & 4.685 {\tiny(0.692)}
& 35.65\% {\tiny(1,174)}  & 3.632 {\tiny(0.480)} & 5.518 {\tiny(0.532)} \\
\cmidrule(lr){1-10}
rus-rest
& 77.69\% {\tiny(4,364)} & 7.248 {\tiny(0.636)} & 6.652 {\tiny(0.842)}
& 4.2\%  {\tiny(236)}  & 4.976 {\tiny(0.238)} & 3.616 {\tiny(1.121)}
& 18.11\% {\tiny(1,017)}  & 2.801 {\tiny(0.677)} & 6.532 {\tiny(0.961)} \\
\cmidrule(lr){1-10}
tat-rest
& 77.69\% {\tiny(4,364)} & 7.248 {\tiny(0.636)} & 6.652 {\tiny(0.842)}
& 4.2\%  {\tiny(236)}  & 4.976 {\tiny(0.238)} & 3.616 {\tiny(1.121)}
& 18.11\% {\tiny(1,017)}  & 2.801 {\tiny(0.677)} & 6.532 {\tiny(0.961)} \\
\cmidrule(lr){1-10}
ukr-rest
& 77.69\% {\tiny(4,364)} & 7.248 {\tiny(0.636)} & 6.652 {\tiny(0.842)}
& 4.2\%  {\tiny(236)}  & 4.976 {\tiny(0.238)} & 3.616 {\tiny(1.121)}
& 18.11\% {\tiny(1,017)}  & 2.801 {\tiny(0.677)} & 6.532 {\tiny(0.961)} \\
\cmidrule(lr){1-10}
zho-rest
& 73.40\% {\tiny(13,295)} & 6.406 {\tiny(0.463)} & 6.016 {\tiny(0.626)}
& 12.47\%  {\tiny(2,258)}  & 4.976 {\tiny(0.354)} & 5.092 {\tiny(0.359)}
& 14.13\% {\tiny(2,559)}  & 3.909 {\tiny(0.400)} & 5.071 {\tiny(0.696)}\\
\cmidrule(lr){1-10}
zho-lap
& 71.04\% {\tiny(9,630)} & 6.405 {\tiny(0.456)} & 5.769 {\tiny(0.674)}
& 8.19\%  {\tiny(1,111)}  & 5.028 {\tiny(0.338)} & 5.185 {\tiny(0.438)}
& 20.78\% {\tiny(2,817)}  & 3.807 {\tiny(0.399)} & 5.402 {\tiny(0.602)}\\
\cmidrule(lr){1-10}
zho-fin
& 65.10\% {\tiny(3,613)} & 6.025 {\tiny(0.302)} & 5.466 {\tiny(0.407)}
& 23.35\%  {\tiny(1,296)}  & 5.222 {\tiny(0.320)} & 5.165 {\tiny(0.281)}
& 11.55\% {\tiny(641)}  & 4.042 {\tiny(0.274)} & 5.326 {\tiny(0.421)} \\

\bottomrule
\end{tabular}
\label{tab:valence_arousal_language_pct}
\end{table}

\subsection{Aspect Sentiment Classification (ASC) Results
}
\begin{table}[H]
\centering
\small
\setlength{\tabcolsep}{2.2pt}
\renewcommand{\arraystretch}{1.0}

\newcommand{\best}[1]{\textbf{#1}}
\begin{tabular*}{\textwidth}{@{\extracolsep{\fill}}ll ccc|ccc|ccc|ccc}
\toprule
&& \multicolumn{6}{c|}{Zero-Shot Learning} 
& \multicolumn{6}{c}{One-Shot Learning} \\
\cmidrule(lr){3-8} \cmidrule(lr){9-14}
Subtask&Dataset & \multicolumn{3}{c|}{\centering GPT-5 mini } & \multicolumn{3}{c|}{Kimi K2 Thinking} & \multicolumn{3}{c|}{\centering GPT-5 mini} & \multicolumn{3}{c}{ Kimi K2 Thinking}\\ 
\cmidrule(lr){3-5}\cmidrule(lr){6-8} \cmidrule(lr){9-11}\cmidrule(lr){12-14}
&& P & R & F1 & P & R & F1 & P & R & F1 & P & R & F1\\ \midrule

\multirow{10}{*}{ASC} & eng-rest & 0.6792 & 0.7045 & 0.6885 & 0.6751 & 0.7104 & \best{0.6902} & 0.6798 & 0.7067 & \best{0.6904} & 0.6692 & 0.7124 & 0.6870\\
&eng-lap & 0.6666 & 0.6723 & \best{0.6626} & 0.6424 & 0.6681 & 0.6501 & 0.6491 & 0.6762 & \best{0.6580} & 0.6370 & 0.6716 & 0.6458\\
&jpn-hot & 0.7027 & 0.7044 & 0.7035 & 0.6972 & 0.7152 & \best{0.7053} & 0.6953 & 0.7045 & 0.6998 & 0.6978 & 0.7183 & \best{0.7069}\\
&jpn-fin & 0.7238 & 0.7233 & \best{0.7235} & 0.6757 & 0.6897 & 0.6792 & 0.7427 &  0.7240 & \best{0.7308} & 0.7113 & 0.7183 & 0.7146\\
&rus-rest & 0.7737 & 0.7283 & \best{0.7452} & 0.7494 & 0.7198 & 0.7295 & 0.7747 & 0.7324 & \best{0.7478} & 0.7426 & 0.7153 & 0.7241\\
&tat-rest & 0.6830 & 0.6755 & 0.6785 & 0.7002 & 0.6973 & \best{0.6975} & 0.6981 & 0.6872 & 0.6921 & 0.6959 & 0.6960 & \best{0.6956}\\
&ukr-rest &0.7320 & 0.7028 & \best{0.7132} & 0.7254 & 0.7047 & 0.7116 & 0.7655 & 0.7286 & \best{0.7427} & 0.7118 & 0.6935 & 0.6995\\
&zho-rest & 0.7622 & 0.7207 & 0.7156 & 0.7630 & 0.7325 & \best{0.7295} & 0.7457 & 0.7177 & 0.7086 & 0.7538 & 0.7329 & \best{0.7249}\\
&zho-lap & 0.7920 & 0.7464 & 0.7540 & 0.7805 & 0.7602 & \best{0.7612} & 0.7803 & 0.7416 & 0.7464 & 0.7897 & 0.7658 & \best{0.7682}\\
&zho-fin&0.7010 & 0.6797 & \best{0.6407} & 0.6303 & 0.6642 & 0.6329 & 0.7010 & 0.6716 & \best{0.6302} & 0.6274 & 0.6619 & 0.6254\\
\cmidrule(lr){1-14}
&AVG & 0.7216 & 0.7058 & \best{0.7025} & 0.7039 & 0.7062 & 0.6987 & 0.7232 & 0.7091 & \best{0.7047} & 0.7037 & 0.7086 & 0.6992\\
\bottomrule
\end{tabular*}

\vspace{0.6em}
\begin{tabular*}{\textwidth}{@{\extracolsep{\fill}}ll ccc|ccc|ccc|ccc}
\toprule
&& \multicolumn{12}{c}{Supervised Fine-Tuning }\\
\cmidrule(lr){3-14}
Subtask&Dataset & \multicolumn{3}{c|}{\centering Qwen3 (14B)} & \multicolumn{3}{c|}{Ministral-3 (14B)} & \multicolumn{3}{c|}{\centering Llama-3.3 (70B)} & \multicolumn{3}{c}{ GPT-OSS (120B)}\\ 
\cmidrule(lr){3-5}\cmidrule(lr){6-8} \cmidrule(lr){9-11}\cmidrule(lr){12-14}
&& P & R & F1 & P & R & F1 & P & R & F1 & P & R & F1\\ \midrule

\multirow{10}{*}{ASC} & eng-rest & 0.6796 & 0.7487 & \best{0.7019} & 0.6121 & 0.6311 & 0.5915 & 0.6140 & 0.6612 & 0.6197 & 0.6739 & 0.7050 & 0.6603\\
&eng-lap & 0.6395 & 0.6807 & 0.6496 & 0.6206 & 0.6014 & 0.5690 & 0.5896 & 0.5922 & 0.5597 & 0.6654 & 0.7066 & \best{0.6548}\\
&jpn-hot & 0.6949 & 0.7732 & \best{0.7080} & 0.6503 & 0.6551 & 0.6431 & 0.6769 & 0.7567 & 0.6892 & 0.7245 & 0.6808 & 0.6928\\
&jpn-fin & 0.6797 & 0.7051 & 0.6768 & 0.6838 & 0.6855 & 0.6831 & 0.6367 & 0.6695 & 0.6402 & 0.7189 & 0.7236 &\best{ 0.7212}\\
&rus-rest & 0.7165 & 0.7204 & 0.7184 & 0.7123 & 0.6605 & 0.6421 & 0.6908 & 0.6934 & 0.6732 & 0.8063 & 0.7320 & \best{0.7549}\\
&tat-rest & 0.5730 & 0.6229 & 0.5915 & 0.5471 & 0.5760 & 0.5060 & 0.4504 & 0.4864 & 0.3712 & 0.7398 & 0.6835 & \best{0.7020}\\
&ukr-rest &0.7226 & 0.7303 & 0.7260 & 0.6640 & 0.6457 & 0.6217 & 0.6882 & 0.6995 & 0.6779 & 0.8008 & 0.7148 & \best{0.7376}\\
&zho-rest & 0.7198 & 0.7166 & \best{0.7133} & 0.7033 & 0.6291 & 0.5614 & 0.6798 & 0.6923 & 0.6718 & 0.7408 & 0.6999 & 0.7060\\
&zho-lap & 0.7465 & 0.7533 & 0.7487 & 0.6989 & 0.5984 & 0.5284 & 0.6802 & 0.6947 & 0.6714 & 0.7676 & 0.7503 & \best{0.7578}\\
&zho-fin&0.6827 & 0.6738 & \best{0.6677} & 0.6206 & 0.6454 & 0.5525 & 0.6011 & 0.6759 & 0.6136 & 0.6478 & 0.6239 & 0.6348\\
\cmidrule(lr){1-14}
&AVG & 0.6855 & 0.7125 & 0.6902 & 0.6513 & 0.6328 & 0.5899 & 0.6308 & 0.6622 & 0.6188 & 0.7286 & 0.7020 & \best{0.7022}\\
\bottomrule
\end{tabular*}
\end{table}

\subsection{ASTE Results}
\begin{table}[H]
\centering
\small
\setlength{\tabcolsep}{2.2pt}
\renewcommand{\arraystretch}{1.0}

\newcommand{\best}[1]{\textbf{#1}}
\begin{tabular*}{\textwidth}{@{\extracolsep{\fill}}ll ccc|ccc|ccc|ccc}
\toprule
&& \multicolumn{6}{c|}{Zero-Shot Learning} 
& \multicolumn{6}{c}{One-Shot Learning} \\
\cmidrule(lr){3-8} \cmidrule(lr){9-14}
Subtask&Dataset & \multicolumn{3}{c|}{\centering GPT-5 mini} & \multicolumn{3}{c|}{Kimi K2 Thinking} & \multicolumn{3}{c|}{\centering GPT-5 mini} & \multicolumn{3}{c}{ Kimi K2 Thinking}\\ 
\cmidrule(lr){3-5}\cmidrule(lr){6-8} \cmidrule(lr){9-11}\cmidrule(lr){12-14}
&& P & R & F1 & P & R & F1 & P & R & F1 & P & R & F1\\ \midrule
\multirow{8}{*}{ASTE} & eng-rest & 0.5654 & 0.6618 & \best{0.6098} & 0.5842 & 0.5932 & 0.5887 & 0.5703 & 0.6501 & \best{0.6076} & 0.5526 & 0.5627 & 0.5576\\
&eng-lap & 0.5141 & 0.6015 & \best{0.5544} & 0.5396 & 0.5554 & 0.5474 & 0.5402 &  0.6395 & \best{0.5857} & 0.4752 & 0.5094 & 0.4917\\
&jpn-hot & 0.2009 & 0.2509 & 0.2231 & 0.3431 & 0.3805 & \best{0.3608} & 0.2728 & 0.3153 & 0.2925 & 0.3827 & 0.3867 & \best{0.3847}\\
&rus-rest & 0.4152 & 0.5435 & 0.4707 & 0.4554 & 0.4908 & \best{0.4724} & 0.3768 & 0.4565 & 0.4128 & 0.4630 & 0.4443 & \best{0.4534}\\
&tat-rest & 0.3526 & 0.4565 & 0.3979 & 0.3858 & 0.4153 & \best{0.4000} & 0.3166 & 0.3840 & 0.3470 & 0.4062 & 0.4084 & \best{0.4073} \\
&ukr-rest & 0.4125 & 0.5275 & \best{0.4630} & 0.4365 & 0.4725 & 0.4538 & 0.3624 & 0.3901 & 0.3757 & 0.4692 & 0.4534 & \best{0.4612}\\
&zho-rest & 0.3009 & 0.4352 & 0.3558 & 0.3668 & 0.4352 & \best{0.3981} & 0.2301 & 0.3114 & 0.2646 & 0.3485 & 0.3827 & \best{0.3648}\\
&zho-lap & 0.2265 & 0.3751 & \best{0.2825} & 0.2452 & 0.3096 & 0.2736 & 0.2385 & 0.3545 & 0.2851 & 0.2588 & 0.3356 & \best{0.2922} \\
\cmidrule(lr){1-14}
&AVG & 0.3735 & 0.4815 & 0.4197 & 0.4196 & 0.4566& \best{0.4369}& 0.3635 & 0.4377 & 0.3964& 0.4195 & 0.4354 & \best{0.4266}\\
\bottomrule

\end{tabular*}
\vspace{0.6em}

\begin{tabular*}{\textwidth}{@{\extracolsep{\fill}}ll ccc|ccc|ccc|ccc}
\toprule
&& \multicolumn{12}{c}{Supervised Fine-Tuning }\\
\cmidrule(lr){3-14}
Subtask&Dataset & \multicolumn{3}{c|}{\centering Qwen3 (14B)} & \multicolumn{3}{c|}{Ministral-3 (14B)} & \multicolumn{3}{c|}{\centering Llama-3.3 (70B)} & \multicolumn{3}{c}{ GPT-OSS (120B)}\\ 
\cmidrule(lr){3-5}\cmidrule(lr){6-8} \cmidrule(lr){9-11}\cmidrule(lr){12-14}
&& P & R & F1 & P & R & F1 & P & R & F1 & P & R & F1\\ \midrule

\multirow{8}{*}{ASTE} & eng-rest & 0.5123 & 0.4871 & 0.4994 & 0.3524 & 0.3593 & 0.3558 & 0.6259 & 0.6059 & \best{0.6158} & 0.5992 & 0.5871 & 0.5931\\
&eng-lap & 0.4302 & 0.4448 & 0.4374 & 0.3486 & 0.3651 & 0.3567 & 0.4972 & 0.4430 & \best{0.4685} & 0.4911 & 0.4446 & 0.4666\\
&jpn-hot & 0.2003 & 0.2010 & 0.2006 & 0.1749 & 0.1968 & 0.1852 & 0.5686 & 0.5627 & \best{0.5657} & 0.5573 & 0.5530 & 0.5551\\
&rus-rest &0.3210 & 0.3496 & 0.3347 & 0.2107 & 0.2229 & 0.2166 & 0.5695 & 0.5565 & \best{0.5629} & 0.5025 & 0.5313 & 0.5165\\
&tat-rest & 0.1744 & 0.1832 & 0.1787 & 0.1414 & 0.1458 & 0.1436 & 0.5276 & 0.4954 & \best{0.5110} & 0.4353 & 0.4626 & 0.4486 \\
&ukr-rest & 0.3161 & 0.3359 & 0.3257 & 0.2004 & 0.2153 & 0.2076 & 0.5655 & 0.5763 & \best{0.5709} & 0.5065 & 0.5321 & 0.5190 \\
&zho-rest & 0.2466 & 0.2587 & 0.2525 & 0.1695 & 0.1943 & 0.1811 & 0.4695 & 0.4708 & \best{0.4702} & 0.4526 & 0.4425 & 0.4475 \\
&zho-lap & 0.1924 & 0.2431 & 0.2148 & 0.1485 & 0.2005 & 0.1706 & 0.4984 & 0.4743 & \best{0.4860} & 0.4386 & 0.4249 & 0.4317 \\
\cmidrule(lr){1-14}
&AVG & 0.2992 & 0.3129 & 0.3055 & 0.2183 & 0.2375 & 0.2272 & 0.5403 & 0.5231 & \best{0.5314} & 0.4979 & 0.4973 & 0.4973
\\
\bottomrule
\end{tabular*}
\end{table}

\subsection{ASQP Results}
\begin{table}[H]
\centering
\small
\setlength{\tabcolsep}{2.2pt}
\renewcommand{\arraystretch}{1.0}

\newcommand{\best}[1]{\textbf{#1}}
\begin{tabular*}{\textwidth}{@{\extracolsep{\fill}}ll ccc|ccc|ccc|ccc}
\toprule
&& \multicolumn{6}{c|}{Zero-Shot learning} 
& \multicolumn{6}{c}{One-Shot Learning} \\
\cmidrule(lr){3-8} \cmidrule(lr){9-14}
Subtask&Dataset & \multicolumn{3}{c|}{\centering GPT-5 mini} & \multicolumn{3}{c|}{Kimi K2 Thinking} & \multicolumn{3}{c|}{\centering GPT-5 mini} & \multicolumn{3}{c}{ Kimi K2 Thinking}\\ 
\cmidrule(lr){3-5}\cmidrule(lr){6-8} \cmidrule(lr){9-11}\cmidrule(lr){12-14}
&& P & R & F1 & P & R & F1 & P & R & F1 & P & R & F1\\ \midrule
\multirow{8}{*}{ASQP} & eng-rest & 0.4446 & 0.5204 & \best{0.4795} & 0.4468 & 0.4537 & 0.4502 & 0.4681 & 0.5336 & \best{0.4987} & 0.4437 & 0.4519 & 0.4478 \\
&eng-lap & 0.2678 & 0.3134 & 0.2888 & 0.3463 & 0.3565 & \best{0.3513} & 0.3272 & 0.3873 & \best{0.3547} & 0.2872 & 0.3078 & 0.2972 \\
&jpn-hotel & 0.1071 & 0.1337 & 0.1190 & 0.1487 & 0.1649 & \best{0.1564} & 0.1667 & 0.1927 & 0.1787 & 0.2188 & 0.2211 & \best{0.2199} \\
&rus-rest & 0.2758 & 0.3611 & 0.3127 & 0.3081 & 0.3321 & \best{0.3196} & 0.2703 & 0.3275 & 0.2962 & 0.3190 & 0.3061 & \best{0.3124}\\
&tat-rest & 0.2205 & 0.2855 & 0.2488 & 0.2752 & 0.2962 & \best{0.2853} & 0.2102 & 0.2550 & 0.2304 & 0.2840 & 0.2855 & \best{0.2847} \\
&ukr-rest & 0.2549 & 0.3260 & 0.2861 & 0.3103 & 0.3359 & \best{0.3226} & 0.2617 & 0.2817 & 0.2713 & 0.3262 & 0.3153 & \best{0.3207} \\
&zho-rest & 0.2369 & 0.3425 & 0.2801 & 0.2858 & 0.3390 & \best{0.3102} & 0.1833 & 0.2482 & 0.2109 & 0.2791 & 0.3065 & \best{0.2922}\\
&zho-lap & 0.1255 & 0.2078 & 0.1565 & 0.1748 & 0.2208 & \best{0.1951} & 0.1640 & 0.2431 & 0.1959 & 0.1999 & 0.2592 & \best{0.2257} \\
\cmidrule(lr){1-14}
&AVG & 0.2416 & 0.3113 & 0.2714 & 0.2870 & 0.3124 & \best{0.2988} & 0.2564 & 0.3086 & 0.2796 & 0.2947 & 0.3067 & \best{0.3001}\\
\cmidrule(lr){1-14}\morecmidrules\cmidrule(lr){1-14}
\end{tabular*}
\vspace{0.6em}

\begin{tabular*}{\textwidth}{@{\extracolsep{\fill}}ll ccc|ccc|ccc|ccc}
\toprule
&& \multicolumn{12}{c}{Supervised Fine-tuning }\\
\cmidrule(lr){3-14}
Subtask&Dataset & \multicolumn{3}{c|}{\centering Qwen3 (14B)} & \multicolumn{3}{c|}{Ministral-3 (14B)} & \multicolumn{3}{c|}{\centering Llama-3.3 (70B)} & \multicolumn{3}{c}{ GPT-OSS (120B)}\\ 
\cmidrule(lr){3-5}\cmidrule(lr){6-8} \cmidrule(lr){9-11}\cmidrule(lr){12-14}
&& P & R & F1 & P & R & F1 & P & R & F1 & P & R & F1\\ \midrule
\multirow{8}{*}{ASQP} & eng-rest & 0.2594 & 0.2466 & 0.2528 & 0.2450 & 0.2499 & 0.2474 & 0.5929 & 0.5740 & \best{0.5833} & 0.5652 & 0.5538 & 0.5594 \\
&eng-lap & 0.1584 & 0.1544 & 0.1564 & 0.1552 & 0.1625 & 0.1588 & 0.3011 & 0.2684 & \best{0.2838} & 0.2925 & 0.2648 & 0.2780 \\
&jpn-hotel & 0.0407 & 0.0381 & 0.0393 & 0.0363 & 0.0409 & 0.0385 & 0.4335 & 0.4290 & \best{0.4312} & 0.4204 & 0.4172 & 0.4188\\
&rus-rest & 0.1650 & 0.1748 & 0.1698 & 0.1162 & 0.1229 & 0.1194 & 0.5227 & 0.5107 & \best{0.5166} & 0.4542 & 0.4802 & 0.4668\\
&tat-rest & 0.0828 & 0.0870 & 0.0849 & 0.0711 & 0.0733 & 0.0722 & 0.4837 & 0.4542 & \best{0.4685} & 0.3958 & 0.4206 & 0.4078 \\
&ukr-rest & 0.1474 & 0.1527 & 0.1500 & 0.1180 & 0.1267 & 0.1222 & 0.5236 & 0.5336 & \best{0.5285} & 0.4629 & 0.4863 & 0.4743\\
&zho-rest & 0.1246 & 0.1307 & 0.1276 & 0.1134 & 0.1300 & 0.1212 & 0.4371 & 0.4383 & \best{0.4377} & 0.4219 & 0.4124 & 0.4171\\
&zho-lap & 0.0995 & 0.1143 & 0.1064 & 0.0927 & 0.1252 & 0.1065 & 0.4001 & 0.3808 & \best{0.3902} & 0.3501 & 0.3392 & 0.3446\\
\cmidrule(lr){1-14}
&AVG & 0.1347 & 0.1373 & 0.1359 & 0.1185 & 0.1289 & 0.1233 & 0.4618 & 0.4486 & \best{0.4550} & 0.4204 & 0.4218 & 0.4209 \\
\cmidrule(lr){1-14}\morecmidrules\cmidrule(lr){1-14}
\end{tabular*}

\end{table}

\end{document}